\documentclass{article}

\usepackage{float}
\usepackage{placeins}
\usepackage{arxiv}
\usepackage[numbers]{natbib}
\usepackage{graphicx}
\usepackage[utf8]{inputenc} %
\usepackage[T1]{fontenc} %
\usepackage{hyperref} %
\usepackage{url} %
\usepackage{booktabs} %
\usepackage{amsfonts} %
\usepackage{nicefrac} %
\usepackage{microtype} %
\usepackage{lipsum}
\usepackage{titlesec}

\usepackage{wrapfig}
\usepackage[font={sf,footnotesize}, labelfont={bf,sf}]{caption}
\usepackage{float}
\usepackage{multirow}
\usepackage{newtxmath}
\usepackage{amsmath}
\usepackage{tabularx}
\usepackage{etoolbox}
\usepackage{subcaption}
\usepackage{comment}
\usepackage{bm}
\usepackage{changepage}
\usepackage{makecell} %
\usepackage{caption}
\usepackage{accents} %

\newcommand{\subcaptiontitle}[1]{
{\sffamily
{\footnotesize
{\centering
\vspace{-10pt}
#1\\
}}
\vspace{4pt}
}}

\DeclareCaptionFormat{myformat}{\fontsize{8}{9}\selectfont#1#2#3}
\newcommand\startsupplement{%
 \makeatletter 
 \renewcommand\thesubsection{\Alph{subsection}.}
 \renewcommand\thesubsubsection{\thesubsection\arabic{subsection}.}

 \setcounter{table}{0}
 \renewcommand{\thetable}{S\arabic{table}}
 \setcounter{figure}{0}
\renewcommand\thefigure{S\arabic{figure}} 
 \makeatother}
 
\newcommand{\addtabletext}[1]{{\setlength{\leftskip}{9pt}\fontsize{8}{10}\selectfont#1}}

\titlespacing*{\section}{0pc}{2ex}{0pt}
\titlespacing*{\subsection}{0pc}{1ex}{-6pt}
\titlespacing*{\subsubsection}{0pc}{0ex}{2pt}

\titleformat{\subsubsection}[runin]%
 {\bfseries}%
 {\thesubsubsection}%
 {0.5em}%
 {}%
 [.]%

\title{Controversial stimuli: pitting neural networks against each other as models of human recognition}

\author{
 Tal Golan\\
 Zuckerman Mind Brain Behavior Institute\\
 Columbia University\\
 New York, NY, USA \\
 \texttt{tal.golan@columbia.edu}
 \And
 Prashant C. Raju \\
 Department of Computer Science\\
 Columbia University\\
 New York, NY, USA\\
 \texttt{prashant.raju@columbia.edu}
 \And
 Nikolaus Kriegeskorte \\
 Zuckerman Mind Brain Behavior Institute,\\ Departments of Psychology, Neuroscience, and Electrical Engineering\\
 Columbia University\\
 New York, NY, USA \\
 \texttt{n.kriegeskorte@columbia.edu}
}

\DeclareMathAlphabet{\pazocal}{OMS}{zplm}{m}{n}
\newcommand{\unif}{\pazocal{U}}

\begin{document}
\maketitle
\begin{abstract} %
Distinct scientific theories can make similar predictions. To adjudicate between theories, we must design experiments for which the theories make distinct predictions. Here we consider the problem of comparing deep neural networks as models of human visual recognition. To efficiently compare models' ability to predict human responses, we synthesize controversial stimuli: images for which different models produce distinct responses. We applied this approach to two visual recognition tasks, handwritten digits (MNIST) and objects in small natural images (CIFAR-10). For each task, we synthesized controversial stimuli to maximize the disagreement among models which employed different architectures and recognition algorithms. Human subjects viewed hundreds of these stimuli, as well as natural examples, and judged the probability of presence of each digit/object category in each image. We quantified how accurately each model predicted the human judgments. The best performing models were a generative Analysis-by-Synthesis model (based on variational autoencoders) for MNIST and a hybrid discriminative-generative Joint Energy Model for CIFAR-10. These DNNs, which model the distribution of images, performed better than purely discriminative DNNs, which learn only to map images to labels. None of the candidate models fully explained the human responses. Controversial stimuli generalize the concept of adversarial examples, obviating the need to assume a ground-truth model. Unlike natural images, controversial stimuli are not constrained to the stimulus distribution models are trained on, thus providing severe out-of-distribution tests that reveal the models' inductive biases. Controversial stimuli therefore provide powerful probes of discrepancies between models and human perception.
\end{abstract} %

Convolutional deep neural networks (DNNs) are currently the best image-computable models of human visual object recognition \cite{kriegeskorte_deep_2015,yamins_using_2016,kietzmann_deep_2019}. To continue improving our computational understanding of biological object recognition, we must efficiently compare different DNN models in terms of their predictions of neuronal and behavioral responses of human and non-human observers. Adjudicating among models requires stimuli for which models make distinct predictions.

Here we consider the problem of adjudicating among models on the basis of their behavior: the classifications of images. Finding stimuli over which high-parametric DNN models disagree is complicated by the flexibility of these models. Given a sufficiently large sample of labeled training images, a wide variety of high-parametric DNNs can learn to predict the human-assigned labels of out-of-sample images. By definition, models with high test accuracy will mostly agree with each other on the classification of test images sampled from the same distribution the training images were sampled from.

Even when there is a considerable difference in test accuracy between two models, the more accurate model is not necessarily more human-like in the features that its decisions are based on. The more accurate model might use discriminative features not used by human observers. DNNs may learn to exploit discriminative features that are completely invisible to human observers \cite{jo_measuring_2017,ilyas_adversarial_2019}. For example, consider a DNN that learns to exploit camera-related artifacts to distinguish between pets and wild animals. Pets are likely to have been photographed by their owners with cellphone cameras and wild animals by professionals with SLR cameras. A DNN that picked up on camera-related features might be similar to humans in its classification behavior on the training distribution (i.e., highly accurate), despite being dissimilar in its mechanism. Another model that does not exploit such features might have lower accuracy, despite being more similar to humans in its mechanism. To reveal the distinct mechanisms, we need to move beyond the training distribution.

There is mounting evidence that even DNN models that exhibit highly human-like responses when tested on in-distribution stimuli often show dramatic deviations from human responses when tested on out-of-distribution (OOD) stimuli.

Prominent examples include images from a different domain \cite[e.g., training a DNN on natural images and testing on silhouettes,][]{kubilius_deep_2016,baker_deep_2018}, as well as images degraded by noise or distortions \cite{dodge_study_2017,geirhos_generalisation_2018,hendrycks_benchmarking_2018}, filtered \cite{jo_measuring_2017}, retextured \cite{geirhos_imagenet-trained_2019}, or adversarially perturbed to bias a DNN's classifications \cite{szegedy_intriguing_2013}. Assessing a model's ability to predict human responses to OOD stimuli provides a severe test of the model's \textit{inductive bias}, i.e., the explicit or implicit assumptions that allow it to generalize from training stimuli to novel stimuli. To correctly predict human responses to novel stimuli, a model has to have an inductive bias similar to that employed by humans. Universal function approximation by itself is insufficient. Previous studies have formally compared the responses of models and humans to distorted \cite{dodge_study_2017,geirhos_generalisation_2018} and adversarially-perturbed images \cite{zhou_humans_2019,elsayed_adversarial_2018}, demonstrating the power of testing for OOD generalization. However, such stimuli are not guaranteed to expose \textit{differences} between different models, because they are not designed to probe the portion of stimulus space where the decisions of different models disagree.

\subsubsection*{Controversial stimuli}

\begin{wrapfigure}[28]{r}{8.7cm}
\vspace{-1mm}
\centering
\includegraphics[width=8.7cm]{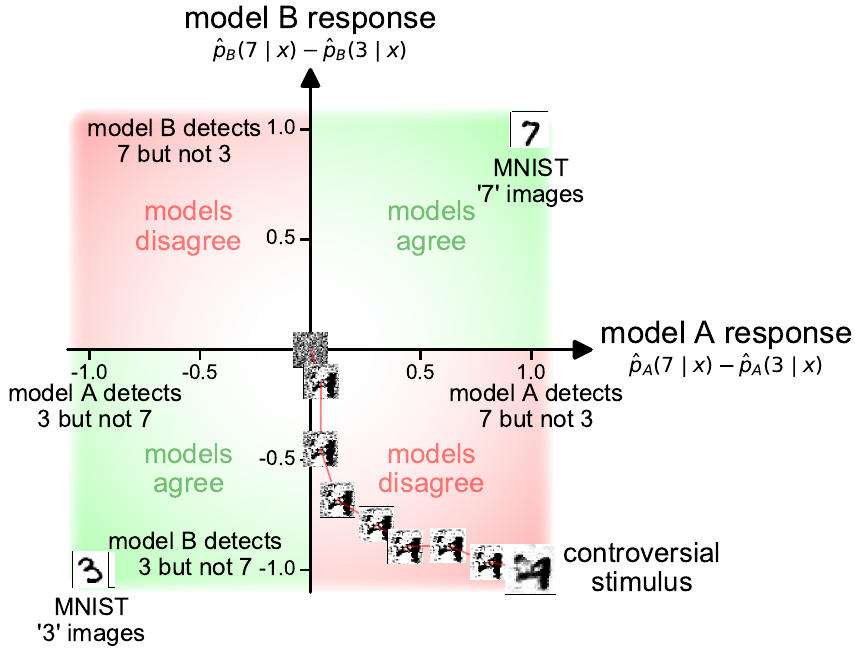}
\caption{\textbf{Synthesizing a single controversial stimulus.} Starting from an initial noise image, one can gradually optimize an image so two (or more) object recognition models disagree on its classification. Here, the resulting controversial stimulus (bottom right) is classified as a 7 by model A and as a 3 by model B. \textbf{Testing such controversial stimuli on human observers allows to determine which of the models has decision boundaries that are more consistent with the human decision boundaries. Often, 'natural' examples (here 50 randomly-selected test MNIST examples) cause no or minimal controversy among models, and therefore lack the power to support efficient comparison of models with respect to human perception.} Model A here is the Capsule Network reconstruction readout, and model B is small VGG$^-$. The stimulus synthesis optimization path (373 steps long) was sampled at nine roughly-equidistant points.}
\label{fig:optimization_path}
\end{wrapfigure}

Here we suggest testing and comparing DNN models of vision on \textit{controversial stimuli}. A controversial stimulus is a sensory input (here, an image) that elicits clearly distinct responses among two or more models. Collecting human responses to stimuli that are controversial between two models gives us great power to adjudicate between the models. The human responses are guaranteed to provide evidence against at least one of the models, since they cannot agree with both models.

Once we define a controversiality score, we can search for such stimuli in large corpora or, more flexibly, synthesize them by optimization (Fig.~\ref{fig:optimization_path}). Stimulus synthesis need not be limited to any particular stimulus prior. If the candidate models differ mostly in how they classify in-distribution examples, an appropriate synthesis procedure, guided by the models' responses, will push the resulting controversial stimuli towards the training distribution. However, if out-of-distribution stimuli evoke considerably different responses among the candidate models, then stimulus synthesis can find them.

\subsubsection*{Controversial stimuli vs. adversarial examples}
Controversial stimuli generalize the notion of adversarial examples. An adversarial example is a stimulus controversial between a model and an oracle that defines the true label. A stimulus that is controversial between two models must be an adversarial example for at least one of them: Since the models disagree, at least one of them must be incorrect (no matter how we choose to define correctness). However, an adversarial example for one of two models may not be controversial between them: both models may be similarly fooled \cite{szegedy_intriguing_2013,goodfellow_explaining_2015,liu_delving_2017}. 
Controversial stimuli provide an attractive alternative to adversarial examples for probing models because they obviate the need for ground-truth labels during stimulus optimization. When adversarially perturbing an image, it is usually assumed that the perturbation will not also affect the true label (in most cases, the class perceived by humans). This assumption necessarily holds only if the perturbation is too small to matter \cite[e.g., as in][]{szegedy_intriguing_2013}. When the bound on the perturbation is large or absent, human observers and the targeted model might actually agree on the content of the image \cite{zhou_humans_2019}, making the image a valid example of another class. Such an image does not constitute a successful adversarial attack. The validity and power of a controversial stimulus, by contrast, are guaranteed given that the stimulus succeeds in making two models disagree.

\subsubsection*{Previous work}
A growing body of literature formally compares DNNs and humans in terms of judgments of natural images \cite{jozwik_deep_2017,rajalingham_large-scale_2018,peterson_evaluating_2018,battleday_capturing_2019,cichy_spatiotemporal_2019,schrimpf_brain-score_2020}. Some of these studies compare different DNNs. However, the field has yet to move toward routine comprehensive inferential comparisons between models that implement alternative theories. Here we systematically and inferentially compare qualitatively distinct models. We introduce a framework in which the models to be tested inform the experimental design, enabling efficient model comparison.

Our approach is conceptually related to Maximum differentiation (MAD) competition \cite{wang_maximum_2008}. MAD competition perturbs a source image in four directions: increasing the response of one model while keeping the response of the other fixed, decreasing the response of one model while keeping the response of the other fixed, and the converse pair (switching the roles of the two models). In contrast, a single controversial stimulus manipulates two (or more) models in opposite directions. Yet crudely speaking, our approach can be viewed as a generalization of MAD competition from univariate response measures (e.g., perceived image quality) to multivariate response measures (e.g., detected object categories) and from local perturbation of natural images to unconstrained search in image space.

\section*{Results}
We demonstrate the approach of controversial stimuli on two relatively simple visual recognition tasks: the classification of hand-written digits \cite[the MNIST dataset,][]{lecun_gradient-based_1998} and the classification of ten basic-level categories in small natural images \cite[the CIFAR-10 dataset,][]{krizhevsky_learning_2009}. From an engineering perspective, both tasks are essentially solved, with multiple, qualitatively different machine learning models attaining near-perfect performance. However, this near-perfect performance on in-distribution examples does not entail that any of the existing models solve MNIST or CIFAR-10 the way humans do. 

\subsubsection*{Synthesizing controversial stimuli}
Consider a set of candidate models. We would like to define a controversiality score for an image $x$. This score should be high if the models strongly disagree on the contents of this image. 

Ideally, we would take an optimal-experimental-design approach \cite{lindley_measure_1956,houlsby_bayesian_2011} and estimate, for a given image, how much seeing the response would reduce our uncertainty about which model generated the data (assuming that one of the models underlies the observed human responses). An image would be preferred according to the expected reduction of the entropy of our posterior belief. However, this statistically-ideal approach is difficult to implement in the context of high-level vision and complex DNN models without relying on strong assumptions.

Here we use a simple heuristic approach. We consider one pair of models ($A$, $B$) at a time. For a given pair of classes, $y_a$ and $y_b$ (e.g., the digits 3 and 7, in the case of MNIST), an image is assigned with a high controversiality score $c_{A,B}^{y_a,y_b}(x)$ if it is recognized by model $A$ as class $y_a$ and by model $B$ as class $y_b$. The following function achieves this:
\begin{equation}\label{eq:C_AB_simplified}
c_{A,B}^{y_a,y_b}(x)=\mathop{\boldsymbol\min}\big\{\hat{p}_A(y_a\mid x),\hat{p}_B(y_b\mid x)\big\}
,
\end{equation} where $\hat{p}_A(y_a\mid x)$ is the estimated conditional probability that image $x$ contains an object of class $y_a$ according to model $A$, and $\mathop{\boldsymbol\min}$ is the minimum function. However, this function assumes that a model cannot simultaneously assign high probabilities to both class $y_a$ and class $y_b$ in the same image. This assumption is true for models with softmax readout. To make the controversiality score compatible also with less restricted (e.g., multi-label sigmoid) readout, we used the following function instead:
\begin{equation}\label{eq:C_AB}
\begin{split}
c_{A,B}^{y_a,y_b}(x)=\mathop{\boldsymbol\min}\big\{\hat{p}_A(y_a\mid x),1-\hat{p}_A(y_b\mid x),\\ \hat{p}_B(y_b\mid x),1-\hat{p}_B(y_a\mid x)\big\}.
\end{split}
\end{equation}

If the models agree over the classification of image $x$, then $\hat{p}_A(y_a\mid x)$ and $\hat{p}_B(y_a\mid x)$ will be either both high or both low, so either $\hat{p}_A(y_a\mid x)$ or $1-\hat{p}_B(y_a\mid x)$ will be a small number, pushing the minimum down.

\begin{wrapfigure}[54]{R}{8.7cm}
\centering
\vspace{-11mm}
\includegraphics[width=8.7cm]{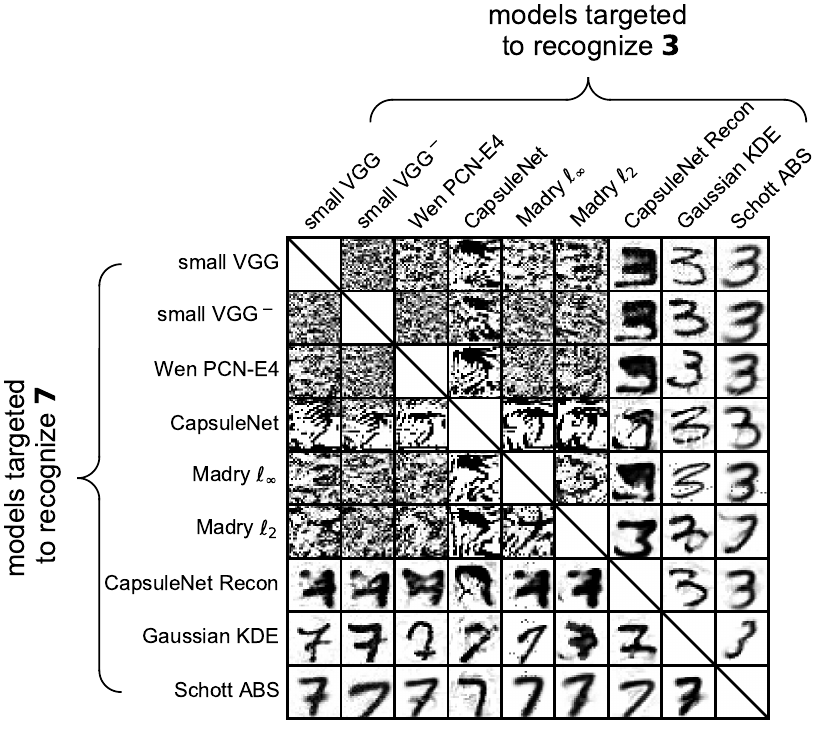}
\caption{\textbf{Synthetic controversial stimuli for one digit pair and all pairs of MNIST models (Experiment 1)}. All these images result from optimizing images to be recognized as containing a 7 (but not a 3) by one model and as containing a 3 (but not a 7) by the other model. Each image was synthesized to target one particular model pair. For example, the bottom-left image (seen as a 7 by human observers) was optimized so that a 7 will be detected with high certainty by the generative ABS model and the discriminative small VGG model will detect a 3. All images here achieved a controversiality score (Eq.~\ref{eq:C_AB}) greater than 0.75.
}
\label{fig:MNIST_controversial_stimuli_particular_digit_pair}
\bigbreak
\includegraphics[width=8.7cm]{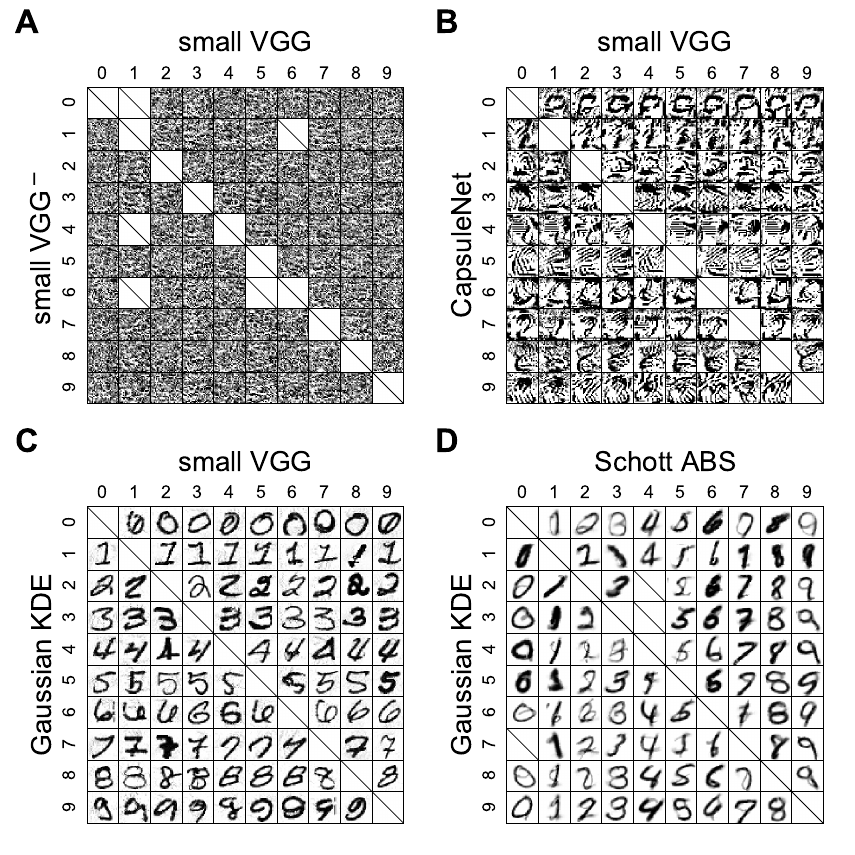}
\caption{\textbf{Synthetic controversial stimuli for all digit pairs and four different MNIST model pairs (Experiment 1)}. The rows and columns within each panel indicate the targeted digits. For example, the top-right image in panel D was optimized so that a 9 (but not a 0) will be detected with high certainty by the Schott ABS model and a 0 (but not a 9) will be detected with high certainty by the Gaussian KDE model. Since this image looks like a 9 to us, it provides evidence in favor of Schott ABS over Gaussian KDE as a model of human digit recognition. Missing (crossed) cells are either along the diagonal (where the two models would agree) or where our optimization procedure did not converge to a sufficiently controversial image (a controversiality score of at least 0.75). See Fig.~\ref*{fig:MNIST_all_controversial_images_by_model_pairs} for all 36 model pairs.}
\label{fig:MNIST_controversial_stimuli_all_digits_four_model_pairs}
\end{wrapfigure}

As in activation-maximization \cite{erhan_visualizing_2009}, we can use gradient ascent to generate images. Here we maximize \eqref{eq:C_AB} by following its gradient with respect to the image (estimated numerically for Experiment 1, and symbolically for Experiment 2).

To increase the efficiency of the optimization and to avoid precision-related issues, the optimization was done on \eqref{eq:C_AB_optimization} (Materials and Methods), a numerically favorable variant of \eqref{eq:C_AB}. We initialized images with uniform white noise and iteratively ascended their controversiality gradient until convergence. A sufficiently controversial resulting image (e.g., $c_{A,B}^{y_a,y_b}(x)\geq0.75$) is not guaranteed. A controversial stimulus cannot be found, for example, if both models associate exactly the same regions of image space with the two classes. However, if a controversial image is found, it is guaranteed to provide a test stimulus for which at least one of the models will make an incorrect prediction.%

\section*{Experiment~1: adjudicating among MNIST models}
\subsubsection*{Candidate MNIST models}
\begin{figure}[ht]
\centering
\includegraphics[width=0.6\textwidth]{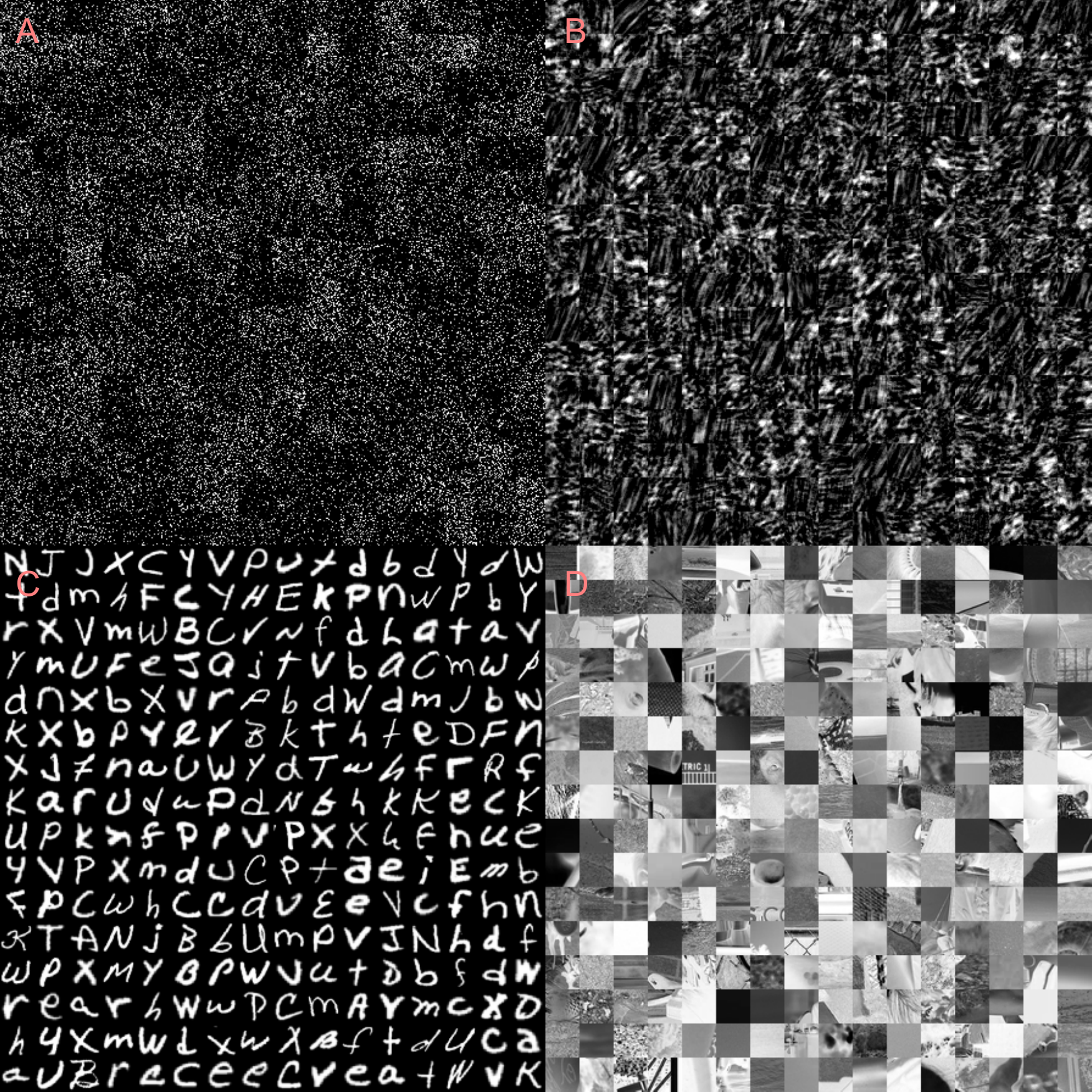}
\caption{
\textbf{A random sample of the negative examples (non-digits) used as a background class for the small VGG${^-}$ model}. (A)~pixel-scrambled MNIST images. (B)~Fourier-phase scrambled MNIST-images. (C)~EMNIST letters \cite{cohen_emnist_2017}, excluding the letters o,s,z,l,i,q and g. (D)~patches cropped from natural images. The small VGG${^-}$ model was trained on the MNIST dataset plus a dataset of a similar size per each of these four non-digit classes (so the digit images were only a fifth of the training set). The non-digit class labels' were coded as all-zero rows in the one-hot coding.
} %
\label{fig:negative_examples}
\end{figure}

We assembled a set of nine candidate models, all trained on MNIST (Table~\ref{tab:MNIST_models_table} and SI section \ref{MNIST model details}). The nine models fall into five families: (1) \textit{Discriminative feedforward models}: an adaptation of the VGG architecture \cite{simonyan_very_2014} to MNIST, trained on either the standard MNIST dataset (`small VGG', see SI subsection \ref{Small VGG}) or on a version extended by non-digit images (`small VGG$^-$', Fig.~\ref*{fig:negative_examples}). (2) \textit{Discriminative recurrent models}: the Capsule Network \cite[][`CapsuleNet']{sabour_dynamic_2017} and the Deep Predictive Coding Network \cite[][`Wen-PCN-E4']{wen_deep_2018}. (3) \textit{Adversarially-trained discriminative models}: DNNs trained on MNIST with either $\ell_\infty$ (`Madry $\ell_\infty$') or $\ell_2$ (`Madry $\ell_2$') norm-bounded perturbations \cite[][]{{madry_towards_2018}}. (4) \textit{A reconstruction-based readout of the Capsule Network} \cite[][`CapsuleNet Recon']{qin_detecting_2020}. (5) \textit{Class-conditional generative models}: models classifying according to a likelihood estimate for each class, obtained from either a class-specific, pixel-space Gaussian Kernel Density Estimator (`Gaussian KDE') or a class-specific Variational Autoencoder (VAE), the `Analysis by Synthesis' model \cite[][`Schott ABS']{schott_towards_2019}.

Many DNN models operate under the assumption that each test image is paired with exactly one correct class (here, an MNIST digit). In contrast, human observers may detect more than one class in an image, or alternatively, detect none. To capture this, the outputs of all of the models were evaluated using multi-label readout, implemented with a sigmoid unit for each class, instead of the usual softmax readout. This setup handles the detection of each class as a binary classification problem \cite{min-ling_zhang_multilabel_2006}.

Another limitation of many DNN models is that they are typically too confident about their classifications \cite{guo_calibration_2017}. To address this issue, we calibrated each model by applying an affine transformation to the preactivations of the sigmoid units (the logits) \cite{guo_calibration_2017}. The slope and intercept parameters of this transformation were shared across classes and were fit to minimize the predictive cross-entropy on MNIST test images. For pre-trained models, this calibration (as well as the usage of sigmoids instead of the softmax readout) affects only the models' certainty and not their classification accuracy (i.e., it does not change the most probable class of each image).

\subsubsection*{Synthetic controversial stimuli reveal deviations between MNIST models and human perception} 
For each pair of models, we formed 90 controversial stimuli, targeting all possible pairs of classes. In Experiment 1, the classes are the ten digits. Fig.~\ref{fig:MNIST_controversial_stimuli_particular_digit_pair} shows the results of this procedure for a particular digit pair across all model pairs. Fig.~\ref{fig:MNIST_controversial_stimuli_all_digits_four_model_pairs} shows the results across all digit pairs for four model pairs.

Viewing the resulting controversial stimuli, it is immediately apparent that pairs of discriminative MNIST models can detect incompatible digits in images that are meaningless to us, the human observers. Images that are confidently classified by DNNs, but unrecognizable to humans are a special type of an adversarial example (described by various terms including `fooling images' \cite{nguyen_deep_2015}, `rubbish class examples' \cite{goodfellow_explaining_2015}, and `distal adversarial examples' \cite{schott_towards_2019}). However, instead of misleading one model (compared to some standard of ground truth), our controversial stimuli elicit disagreement between \textit{two} models. For pairs of discriminatively trained models (Fig.~\ref{fig:MNIST_controversial_stimuli_all_digits_four_model_pairs}A, B), human classifications are not consistent with either model, providing evidence against both.

One may hypothesize that the poor behavior of discriminative models when presented with images falling into none of the classes results from the lack of training on such examples. %
However, the small VGG$^-$ model, trained with diverse non-digit examples, still detected digits in controversial images that are unrecognizable to us (Fig.~\ref{fig:MNIST_controversial_stimuli_all_digits_four_model_pairs}A). %

There were some qualitative differences among the stimuli resulting from targeting pairs of discriminative models. Images targeting one of the two discriminative recurrent DNN models, the Capsule network \cite{sabour_dynamic_2017} and the PCN \cite{wen_deep_2018}, showed increased (yet largely humanly unrecognizable) structure (e.g., Fig.~\ref{fig:MNIST_controversial_stimuli_all_digits_four_model_pairs}B). When the discriminative models pitted against each other included a DNN that had undergone $\ell_2$\nobreakdash-bounded adversarial training \cite{madry_towards_2018}, the resulting controversial stimuli showed traces of human-recognizable digits (Fig.~\ref{fig:MNIST_controversial_stimuli_particular_digit_pair}, Madry $\ell_2$). These digits' human classifications tended to be consistent with the classifications of the adversarially trained discriminative model \cite[see][for a discussion of  $\ell_2$ adversarial training and perception]{tsipras_robustness_2019}.

And yet, when any of the discriminative models was pitted against either the reconstruction-based readout of the Capsule Network, or either of the generative models (Gaussian KDE or ABS), the controversial image was almost always a human-recognizable digit consistent with the target of the reconstruction-based or generative model (e.g., Fig.~\ref{fig:MNIST_controversial_stimuli_all_digits_four_model_pairs}C). Finally, synthesizing controversial stimuli to adjudicate between the three reconstruction-based/generative models produced images whose human classifications are most similar to the targets of the ABS model (e.g., Fig.~\ref{fig:MNIST_controversial_stimuli_all_digits_four_model_pairs}D).

\begin{figure}
\centering
\includegraphics[width=4.49in]{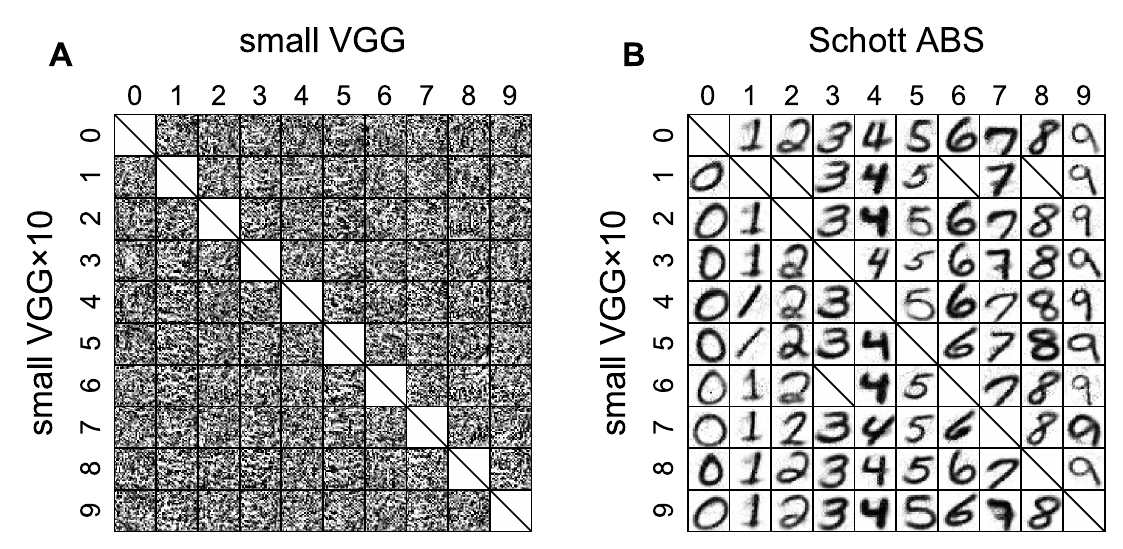}
\caption{\textbf{A set of separate digit-specific discriminative DNNs do not match the set of digit-specific VAEs (the Schott ABS model) in human consistency.} One potential alternative explanation of the advantage of the Schott ABS model is that the ABS model has one DNN per MNIST class, whereas the other candidate MNIST models we tested have a single DNN that learns to detect each of the ten classes. To test this alternative explanation, we trained a variant of the small VGG model where ten small VGG DNNs were independently trained on binary one-vs-all classification tasks (i.e., the first DNN is trained to detect the digit 0 vs. the other digits, the second DNN is trained to detect the digit 1 vs. the other digits, and so on). Other than this modification, the training procedure of each digit-specific DNN was identical to the small-VGG training procedure specified in SI subsection A.1. The scalar sigmoid outputs of the ten DNNs, each representing the detection of one particular digit, were concatenated to form a joint ten-unit output layer. The resulting model ('small VGG$\times$10') had MNIST test error of 0.64\%, slightly worse than the single DNN small VGG model (0.47\%). We synthesized controversial stimuli targeting this model vs. the single DNN small VGG (panel A), as well as this model vs. the Schott ABS model (panel B). \textbf{As can be seen above, there is no qualitative indication that having one discriminative DNN per class improves the model's human consistency.} One can form controversial stimuli for the single-DNN small VGG model and the small VGG$\times$10 model that do not resemble digits (panel A). Furthermore, the human recognition of the controversial stimuli for the small VGG$\times$10 and the Schott ABS models is completely aligned with the Schott ABS model (panel B). In fact, we hypothesize that not sharing parameters between classes is a shortcoming rather than a strength of the Schott ABS model, since parameter sharing enables reusing visual features across different classes.}
\label{fig:duplicated_VGG_control}
\end{figure}

\begin{figure}
\centering
\begin{subfigure}[t]{2.40736in}
\caption{}
\vspace{1.400in}
\includegraphics[width=2.40736in]{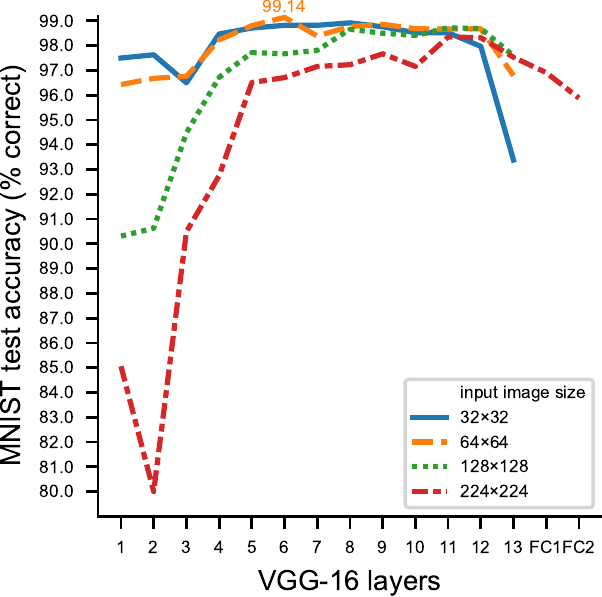}
\end{subfigure}\hfill%
\begin{subfigure}[t]{4.03056in}
\caption{}
\includegraphics[width=4.03056in]{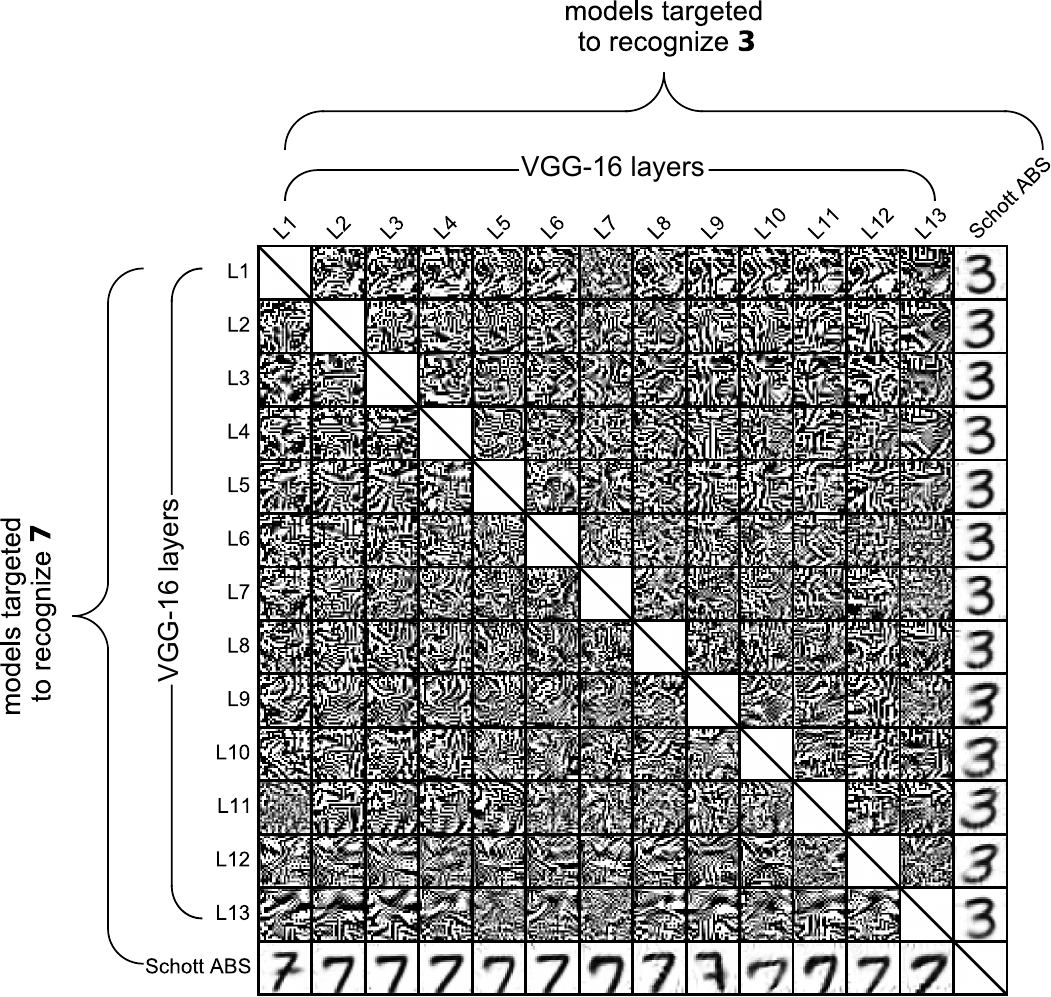}
\end{subfigure}\hfill%

\caption{\textbf{MNIST linear classifiers reading out hidden-layer features of an ImageNet-trained VGG-16 do not exhibit human-consistent classification.} Discriminative DNNs trained on object recognition in natural images (e.g., the ImageNet dataset) were found to have hidden-layer features that can serve as a surprisingly good basis for predicting neural responses in mid-level ventral visual stream \cite[e.g.,][]{yamins_performance-optimized_2014,khaligh-razavi_deep_2014,guclu_deep_2015,cichy_comparison_2016,eickenberg_seeing_2017,wen_neural_2017}, as well as for predicting behavioral similarity judgments \cite[e.g.,][]{peterson_evaluating_2018,zhang_unreasonable_2018}. These findings suggest an alternative explanation of the human-inconsistency of the discriminative models in Experiment 1 (e.g., the small VGG model): perhaps discriminative models can perform well in the MNIST controversial stimuli benchmark if they are first trained on recognizing objects in natural images, and only then learn to classify MNIST as a transfer learning task. After all, the 'visual diet' humans are exposed to from birth consists mostly of natural images. To test for this hypothesis, we trained multi-label linear classifiers on the MNIST task using the activations of the convolutional layers of an Imagenet-trained VGG-16 \cite{simonyan_very_2014} as fixed features. \\(A) \textbf{MNIST test accuracy of multi-label linear classifiers, each operating on the activations of one VGG-16 convolutional layer as a fixed feature set.} Each classifier is trained to classify MNIST digits from the activations of a single layer. Different lines represent classifiers with different degrees of input-image upscaling. Convolutional layers can be evaluated with input images of arbitrary size (as long as they are big enough to fit the kernels), so our rescaled input images were not embedded in blank margins. Fully connected layers (which do not share this flexibility) were tested only with the largest input-image scale (224$\times$224 pixels), on which the model was originally trained. Following this accuracy comparison between different input-image scales and layers, we chose for further testing (panel B) the classifiers that use VGG-16's convolutional layers (i.e., layers 1-13) with a 64$\times$64 pixels input-image size (orange dashed line in panel A). \\(B) \textbf{Controversial stimuli (targeting '3' vs. '7' classification), pitting the linear classifiers trained on the features of the different VGG-16 convolutional layers against each other or against the Schott ABS model. The greater human-consistency of the Schott ABS model is evident, even though this model does not enjoy the advantage of using Imagenet-driven features.} Furthermore, controversial stimuli for pairs of classifiers based on different VGG-16 layers do not resemble human-recognizable digits.\\\textbf{While we cannot preclude that features gained from learning to recognize objects in natural images might be necessary for achieving human-consistent responses in the MNIST task, the controversial stimuli above indicate that in the discriminative-training context, using such features is an insufficient condition for model-human consistency.}%
\\Classifier training details: We used a pretrained VGG-16 model (Keras implementation) and prepended a bilinear upsampling operation to the model's normalized input, transforming MNIST images from the dimensionality of 28$\times$28$\times$1 to 32$\times$32$\times$3, 64$\times$64$\times$3, 128$\times$128$\times$3 or 224$\times$224$\times$3. We then trained an MNIST linear multi-label classifier separately for each layer using the layer's ReLU activations as fixed features. Linear classification was implemented as a new fully-connected layer projecting the ReLU activations of one particular hidden-layer to ten units. A sigmoid activation function was then applied to each of the ten units, rendering this setup equivalent to one-vs-all logistic regression. We minimized the cross-entropy of the sigmoids and a one-hot label representation using Adam \cite{kingma_adam_2014} (Keras implementation) with a learning rate of 0.001 reduced by a factor of 0.2 after every three epochs of no training loss reduction, down to a minimum learning rate of $10^{-6}$. Early stopping was applied based on held-out validation data.}
\label{fig:MNIST_VGG16_layers_LR_control}
\end{figure}

\begin{figure}
\makebox[\textwidth][c]{%
\begin{minipage}{7in}
\centering
\begin{subfigure}[t]{3.0825in}
\captionsetup{margin=0.0in}
\caption{}
\subcaptiontitle{Initializing with 3s}
\includegraphics[height=2.735in]{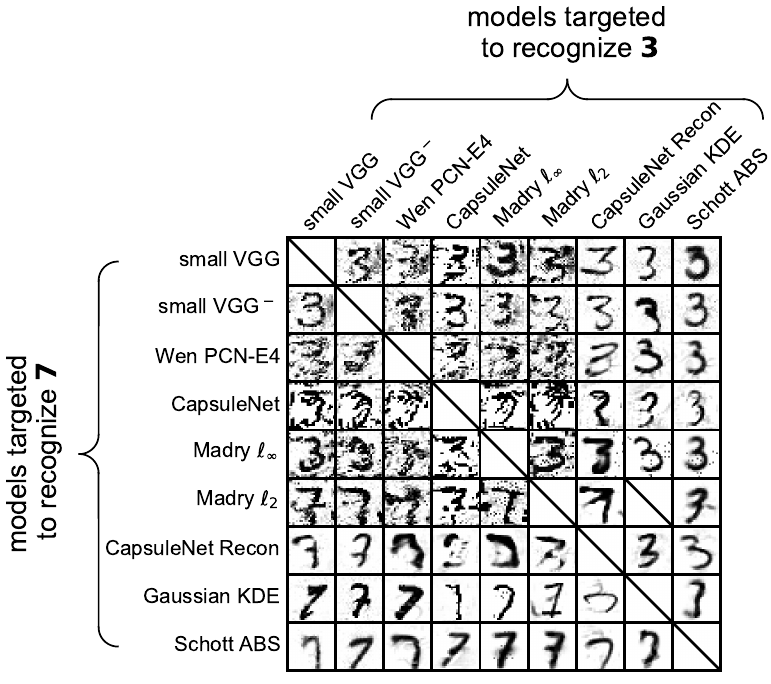}
\end{subfigure}\hfill%
\begin{subfigure}[t]{1.9425in}
\captionsetup{margin=0.01in}
\caption{}
\subcaptiontitle{Initializing with 7s}
\includegraphics[height=2.735in]{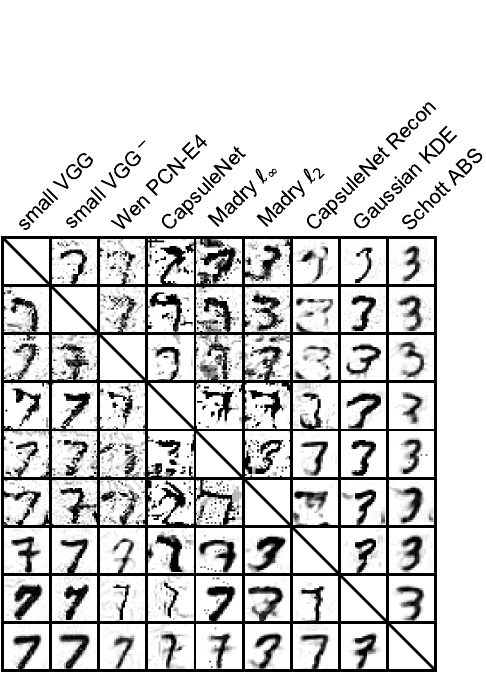}
\end{subfigure}%
\begin{subfigure}[t]{1.9425in}
\captionsetup{margin=0.01in}
\caption{}
\subcaptiontitle{Initializing with 8s}
\includegraphics[height=2.735in]{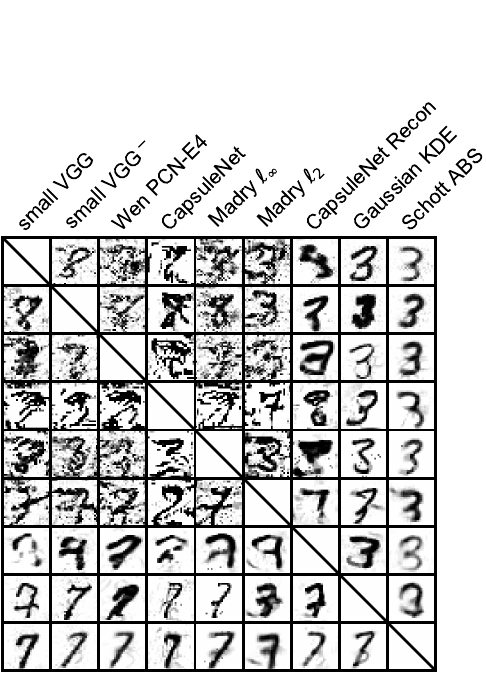}
\end{subfigure}
\end{minipage}}
\caption{\textbf{Initializing controversial stimuli with digit images rather than with random-noise images does not change the observed hierarchy of human-consistency among the tested MNIST models.} Each controversial stimulus was initialized with an MNIST test example randomly selected from either the 3 (A), 7 (B) or 8 (C) class. These images were then optimized using the same procedure employed in Experiment 1 to evoke a 3 (but not 7) detection in the models listed above each panel and at the same time, a 7 (but not 3) detection in the models listed to the left of each panel.\textbf{The greater alignment of the Schott ABS \cite{schott_towards_2019} model with human perception compared to the other candidate models is reproduced for each of the three different initializations: Most of the images in the rightmost column of each panel look like the digit 3, a judgment consistent with the target class of the ABS model for these images. Similarly, most of the images in the bottom row of each panel look like the digit 7, again a human perceptual judgment consistent with the target class of the ABS model and inconsistent with the other models.}
While this analysis is provided here as a control for the effect of initialization, we generally recommend initializing controversial stimuli with random-noise images. Controversial stimuli optimized from non-random initial images may inherit human discernible features from the seed image. The resulting controversial stimuli may thus contain natural features, which are not driven by any of the targeted models (as in the top left part of each panel above, where standard discriminative models are paired). In contrast, when a controversial stimulus is initialized with a random noise image, any discernible feature is model-driven rather than experimenter-driven, eliminating the potential confounding factor of perceivable visual content originating from the initializing image. Furthermore, while there is no evidence that gradient obfuscation \cite{athalye_obfuscated_2018} plays a role in the optimization of the controversial stimuli above, initializing controversial stimuli with random images provides a strong protection against this issue. A model with obfuscated gradients will not be able to drive a random noise image towards a high-confidence detection of its target class. Such an outcome transparently indicates an optimization issue rather than give a false impression of model robustness.}
\label{fig:non_random_initialization}
\end{figure}

The ABS model is unique in having one DNN per class, raising the question of whether this, rather than its generative nature, explains its performance. However, imitating this structure by training ten small VGG models as ten binary classifiers did not increase the human consistency of the small VGG model (Fig.~\ref{fig:duplicated_VGG_control}). Another possibility is that a higher-capacity discriminative model with more human-like visual training on natural images might perform better. However, MNIST classification using visual features extracted from the hidden layers of an Imagenet-trained VGG-16 did not outperform the ABS model (Fig.~\ref{fig:MNIST_VGG16_layers_LR_control}). Finally, the advantage of the ABS model persisted also when the optimization was initialized from MNIST test examples instead of random noise images (Fig.~\ref{fig:non_random_initialization}).

\subsubsection*{Human psychophysics can formally adjudicate among models and reveal their limitations}
Inspecting a matrix of controversial stimuli synthesized to cause disagreement among two models can provide a sense of which model is more similar to us in its decision boundaries. However, it does not tell us how a third, untargeted model responds to these images. Moreover, some of the resulting controversial stimuli are ambiguous to human observers. We therefore need careful human behavioral experiments to adjudicate among models.

We evaluated each model by comparing its judgments to those of human subjects and compared the models in terms of how well they could predict the human judgments. For Experiment 1, we selected 720 controversial stimuli (20 per model-pair comparison, see SI section \ref{controversial stimulus selection}) as well as 100 randomly selected MNIST test images. We presented these 820 stimuli to 30 human observers, in a different random order for each observer. For each image, observers rated each digit's probability of presence from 0\% to 100\% on a five-point scale (Fig.~\ref*{fig:Experiment_1_human_exp_demo_trial}). The probabilities were not constrained to sum to 1, so subjects could assign high probability to multiple digits or zero probability to all of them for a given image. There was no objective reference for correctness of the judgments, and no feedback was provided.

For each human subject $s_i$ and model $M$, we estimated the Pearson linear correlation coefficient between the human and model responses across stimuli and classes:
\begin{equation}\label{eq:r_Pearson}
r(M,s_i)=\frac{
\sum\limits_{x,y}\big(\hat{p}_{s_i}(y\mid x) - \bar{\hat{p}}_{s_i}\big)
\big(\hat{p}_M(y \mid x) - \bar{\hat{p}}_M\big)}
{
\sqrt{\sum\limits_{x,y}\big(\hat{p}_{s_i}(y\mid x) - \bar{\hat{p}}_{s_i}\big)^2}
\sqrt{
\sum\limits_{x,y}\big(\hat{p}_M(y \mid x) - \bar{\hat{p}}_M\big)^2}
},
\end{equation} where $\hat{p}_{s_i}(y \mid x)$ is the human-judged probability that image $x$ contains class $y$, $\hat{p}_M(y \mid x)$ is the model's corresponding judgment, $\bar{\hat{p}}_{s_i}$ is the mean probability judgement of subject $s_i$ and $\bar{\hat{p}}_M$ is the mean probability judgment of the model. The overall score of each model was set to its mean correlation coefficient, averaged across all subjects: $\bar{r}_M=\frac{1}{n}\sum\limits_i r(M,s_i)$, where $n$ is the number of subjects.

\begin{figure}[h]
\makebox[\textwidth][c]{%
\begin{minipage}[c]{7in}
\centering
\begin{subfigure}[t]{8.8cm}
\caption{}
\includegraphics[width=8.8cm]{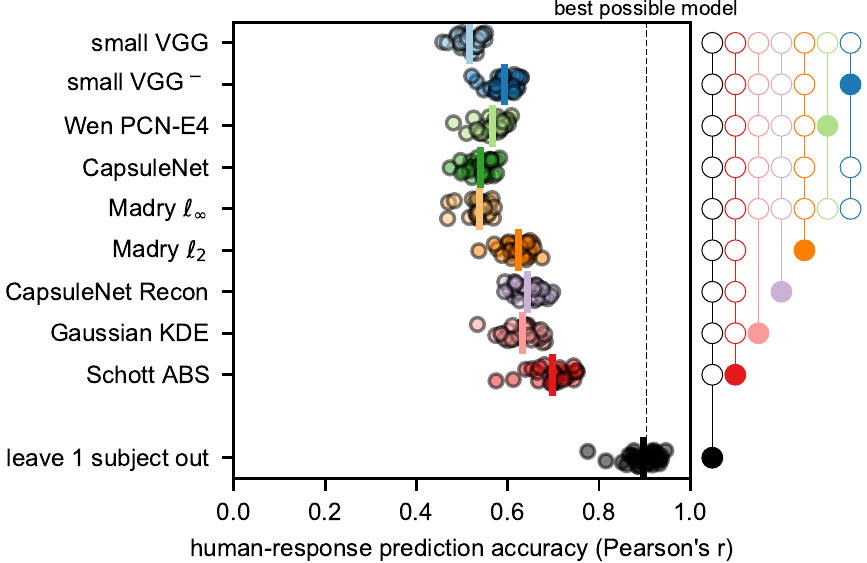}
\label{fig:MNIST_human_responses_prediction_mean_correlation}
\end{subfigure}\hfill%
\begin{subfigure}[t]{8.8cm}
\caption{}
\includegraphics[width=8.8cm]{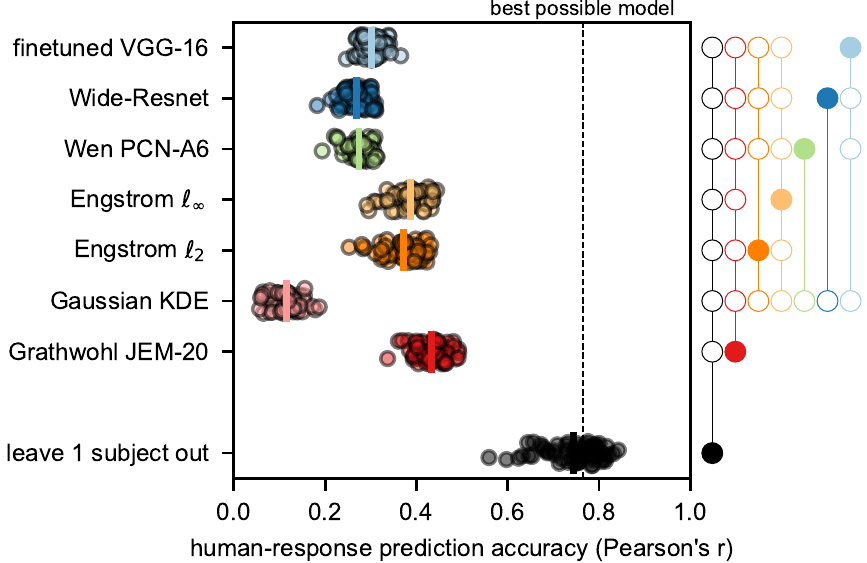}
\label{fig:CIFAR-10_human_responses_prediction_mean_correlation}
\end{subfigure}%
\end{minipage}}
\caption{\textbf{The performance of the candidate MNIST (A) and CIFAR-10 (B) models in predicting the human responses to the entire stimulus set.} Each dot marks the correlation coefficient between the responses of one individual human participant and one model (Eq. \ref{eq:r_Pearson}). The vertical bars mark across-subject means ($\bar{r}_{M}$). The gray dots mark the correlation between each participant's responses and the mean response pattern of the other participants. The mean of the gray dots (a black bar) is the lower bound of the noise ceiling. The dashed line (`best possible model') marks the highest across-subject mean correlation achievable by any single model (upper bound of the noise ceiling). Significance indicators (on the right of each panel): A closed dot connected to a set of open dots indicates that the model aligned with the closed dot has significantly higher correlation than any of the models aligned with the open dots (\textit{p} $<$ 0.05, subject-stimulus bootstrap). Testing controlled the family-wise error rate at 0.05, accounting for the total number of model-pair comparisons (45 for Experiment 1, 28 for Experiment 2). For equivalent analyses with alternative measures of human-response prediction accuracy, see Fig.~\ref{fig:alternative_model_human_prediction_accuracy_measures}. \textbf{The deep generative model (ABS, Experiment 1) and the deep hybrid model (JEM-20, Experiment 2) (both in red) explain human responses to the combined set of controversial and natural stimuli better than all of the other candidate models. And yet, none of the models account for all explainable variance: predicting each subject from her/his peers' mean response pattern achieves significantly higher accuracy.}}%
\label{fig:human_responses_prediction_mean_correlation}
\end{figure}

Given the intersubject variability and decision noise, the true model (if it were included in our set) cannot perfectly predict the human judgments. We estimated a lower bound and an upper bound on the maximal attainable performance (the noise ceiling, see SI subsection \ref{noise_ceiling_details}). The lower bound of the noise ceiling ('leave-one-subject-out', black bars in Fig.~\ref{fig:human_responses_prediction_mean_correlation}A-B) was estimated as the mean across subjects of the correlation between each subject's response pattern and the mean response pattern of the other subjects \cite{nili_toolbox_2014}. The upper bound of the noise ceiling (`best possible model', dashed lines in Fig.~\ref{fig:human_responses_prediction_mean_correlation}A-B) is the highest across-subject-mean correlation achievable by any possible set of predictions.

The results of Experiment 1 (Fig.~\ref{fig:MNIST_human_responses_prediction_mean_correlation}) largely corroborate the qualitative impressions of the controversial stimuli, indicating that the deep class-generative ABS model \cite{schott_towards_2019} is superior to the other models in predicting the human responses to the stimulus set. Its performance is followed by the Gaussian KDE, the reconstruction-based readout of the Capsule network, and the Madry $\ell_2$ adversarially trained model. The other models (all discriminative) performed significantly worse. All models were significantly below the lower bound of the noise ceiling (the black bar in Fig.~\ref{fig:MNIST_human_responses_prediction_mean_correlation}), indicating that none of the models fully explained the explainable variability in the data.

We also evaluated the models separately for controversial stimuli and natural stimuli (i.e., MNIST test images, Fig.~\ref{fig:MNIST_human_responses_prediction_mean_correlation_only_natural}). The ABS and Gaussian KDE models were not as good as the discriminative models in predicting the human responses to the natural MNIST test images, indicating that the discriminative models are better at achieving human-like responses within the MNIST training distribution.

\newpage

\section*{Experiment 2: adjudicating among CIFAR-10 models}

The MNIST task has two obvious disadvantages as a test case: (a) its simplicity compared to visual object recognition in natural images, and (b) the special status of handwritten characters, which are generated through human movement. In Experiment 2, we applied the method of controversial stimuli to a set of models designed to classify small natural images from the CIFAR-10 image set. The purely generative ABS model is reported to fail to scale up to CIFAR-10 \cite{schott_towards_2019}. We therefore included the Joint Energy Model \cite[JEM,][]{grathwohl_your_2019}, which implements a hybrid discriminative-generative approach to CIFAR-10 classification.

\subsubsection*{Candidate CIFAR-10 models}
We assembled a set of seven CIFAR-10 candidate models (Table~\ref{tab:CIFAR10_models_table} and SI section \ref{CIFAR-10 model details}). The seven models fall into five model families largely overlapping with the model families tested in Experiment 1: (1) \textit{Discriminative feedforward models}: a VGG-16 \cite{simonyan_very_2014} first trained on ImageNet and then retrained on upscaled CIFAR-10 (`finetuned VGG-16') and a Wide-Resnet trained exclusively on CIFAR-10 \cite[]['Wide-Resnet']{zagoruyko_wide_2016}. (2) \textit{A discriminative recurrent model}: a CIFAR-10 variant of the Deep Predictive Coding Network \cite[][`Wen-PCN-A6']{wen_deep_2018}. (3) \textit{Adversarially-trained discriminative models}: Resnet-50 DNNs trained on CIFAR-10 with either $\ell_\infty$ (`Engstrom $\ell_\infty$') or $\ell_2$ (`Engstrom $\ell_2$') norm-bounded perturbations \cite[][]{engstrom_robustness_2019}. (4) \textit{A class-conditional generative model}: the pixel-space Gaussian Kernel Density Estimator (`Gaussian KDE'). (5) \textit{A hybrid discriminative-generative models}: the Joint Energy Model \cite[][`Grathwol JEM-20']{grathwohl_your_2019}.

The hybrid JEM has the same WRN-28-10 architecture \cite{zagoruyko_wide_2016} as the discriminative Wide-Resnet model mentioned above, but its training combines a discriminative training objective (minimizing the classification error) with a generative training objective. The generative objective treats the LogSumExp of the DNN's logits as an unnormalized image likelihood estimate and encourages high likelihood assignments to in-distribution images. Including the generative objective in the training improves the model's robustness to adversarial attacks \cite{grathwohl_your_2019}. The model's robustness can be further improved by refining the input-layer representation during inference, nudging it to have higher likelihood. We have tested the JEM model with 20 refinement steps (hence we refer to it here as `JEM-20').

\begin{wrapfigure}[29]{R}{8.7cm}
\centering
\includegraphics[width=8.7cm]{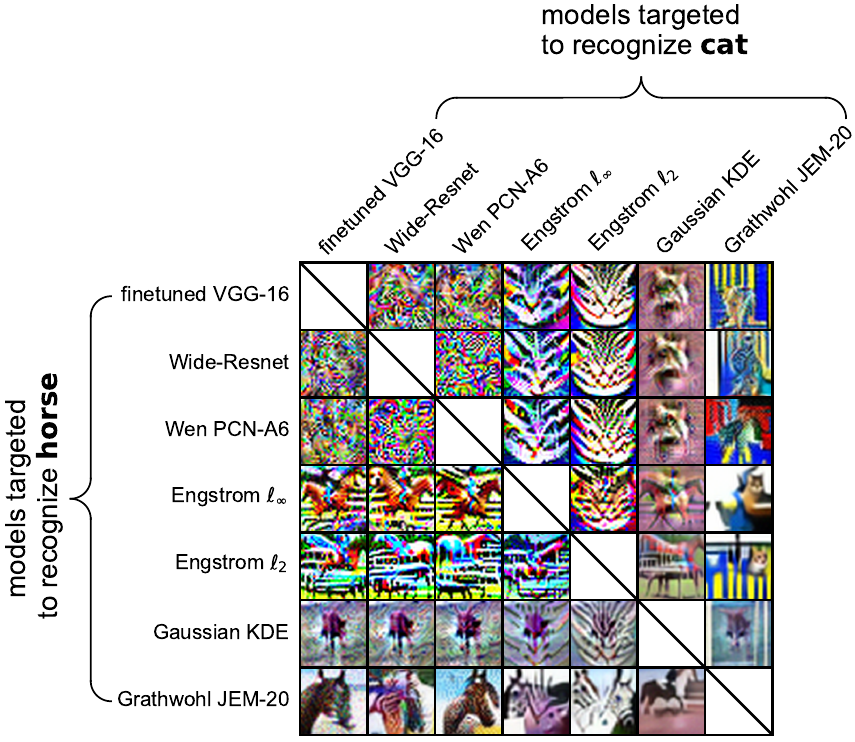}
\caption{\textbf{Synthetic controversial stimuli contrasting the seven different CIFAR-10-classifying models.} Each stimulus results from optimizing an image to be detected as a cat (but not as a horse) by one model and as a horse (but not as a cat) by another model. For example, the bottom-left image (seen as a horse by human observers) was optimized so that the hybrid discriminative-generative JEM-20 model will detect a horse and the discriminative, finetuned VGG-16 model will detect a cat. All images here achieved a controversiality score (Eq.~\ref{eq:C_AB}) greater than 0.75. The images are shown in an upsampled format as presented to the human subjects. See Fig.~\ref{fig:CIFAR10_all_controversial_images_by_model_pairs} for all class combinations.
}
\label{fig:CIFAR_10_controversial_stimuli_particular_class_pair}
\end{wrapfigure}
As in Experiment 1, we used sigmoid readout to allow for more flexible responses, such as detecting multiple or none of the categories. Since the candidate models had a wide range of test accuracies (Table \ref{tab:CIFAR10_models_table}), the sigmoid readout was calibrated for each model such that negative examples would be assigned a median probability of 0.1 and positive examples a median probability of 0.9.

\subsubsection*{Synthetic controversial stimuli reveal deviations between CIFAR-10 models and human perception} 
\begin{figure}
\makebox[\textwidth][c]{
\includegraphics[width=7in]{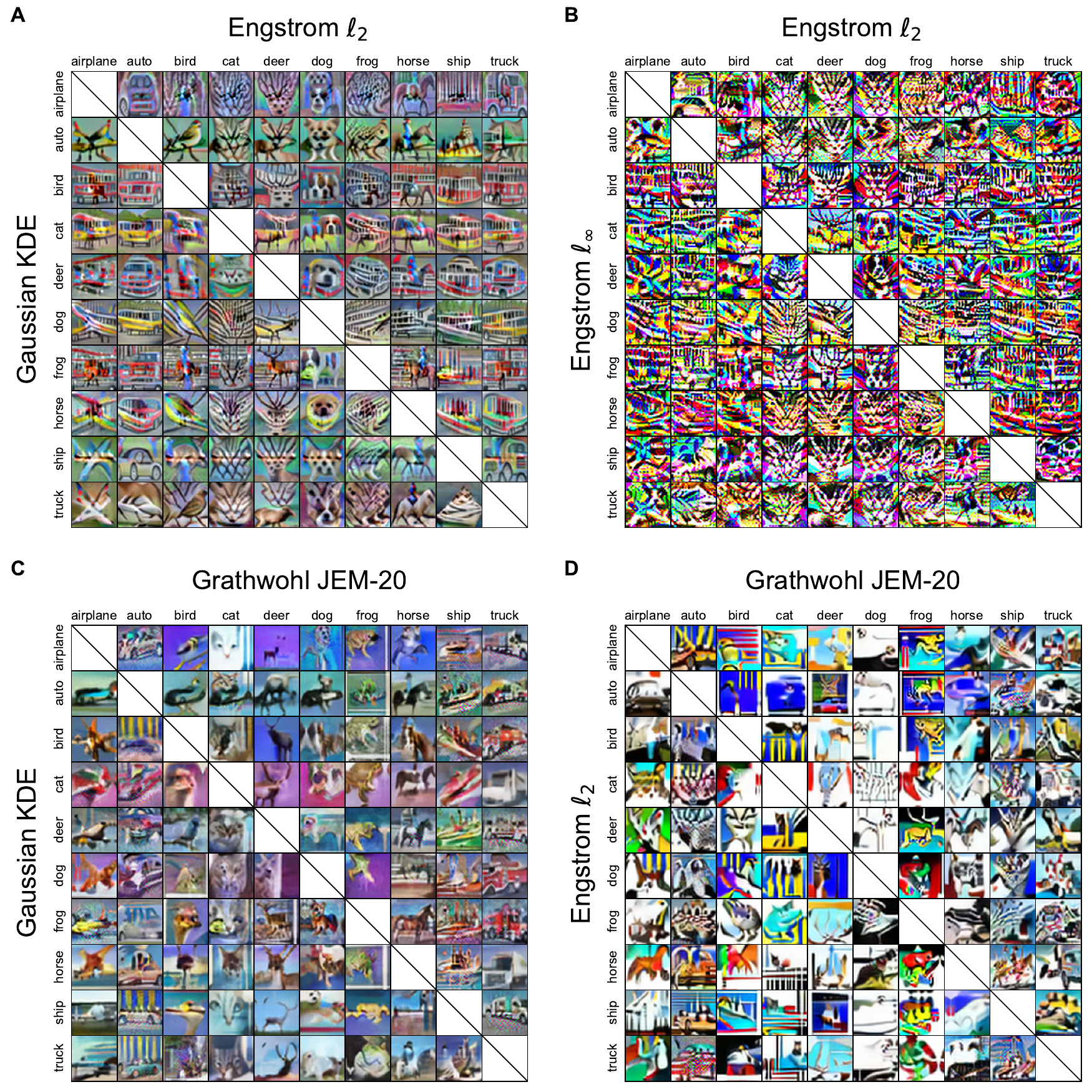}}
\caption{\textbf{Controversial stimuli for CIFAR-10 models, here organized to show the results of targeting each possible CIFAR class pair for four different model pairs}. The rows and columns within each panel indicate the targeted classes. For example, the top-right image in panel C was optimized to be detected as a truck by the JEM-20 model and as an airplane by the Gaussian KDE model. Missing (crossed) cells are along the diagonal, where the two models would agree. All images here have a controversiality score of at least 0.75. See Fig.~\ref*{fig:CIFAR10_all_controversial_images_by_model_pairs} for all 21 model pairs.}
\label{fig:CIFAR_10_controversial_stimuli_all_digits_four_model_pairs}
\end{figure}

Examples of the resulting controversial stimuli appear in Figs.~\ref{fig:CIFAR_10_controversial_stimuli_particular_class_pair} and \ref{fig:CIFAR_10_controversial_stimuli_all_digits_four_model_pairs}. When DNNs trained with a non-adversarial discriminative objective (i.e, the finetuned VGG-16, the discriminatively trained Wide-Resnet and the Predictive Coding Network) are paired with each other, the resulting controversial stimuli do not appear to humans to contain objects of any of the categories. These results bear strong resemblance to those in Experiment 1. In contrast to Experiment 1, however, the target categories for the Gaussian KDE were, by and large, not discernible to humans, indicating that this shallow-generative model, which worked surprisingly well on MNIST, does not scale up to CIFAR-10. Pitting the Gaussian KDE against the JEM-20 model (Fig.~\ref{fig:CIFAR_10_controversial_stimuli_all_digits_four_model_pairs}C) produced almost naturally looking images, in which the target categories of JEM-20 are discernible. In some of these images, low-level features suggestive of the target category of the Gaussian KDE can also be recognized. Also, the target categories of the adversarially trained models were more discernible than in Experiment 1 (Fig.~\ref{fig:CIFAR_10_controversial_stimuli_all_digits_four_model_pairs}A, B). Finally, Pitting the JEM-20 model against one of the adversarially trained models (Fig.~\ref{fig:CIFAR_10_controversial_stimuli_all_digits_four_model_pairs}D) often produced images in which the target category for JEM-20 was discernible. In some images, however, the human-perceptible category was the target of the adversarially trained DNN, or, both or neither of the categories were perceptible. These ambiguities suggest deviations of both JEM-20 and the adversarially trained DNNs from human perception and emphasize the importance of quantitative behavioral experiments.

\begin{figure}[tbhp]
\makebox[\textwidth][c]{%
\begin{minipage}[c]{7in}
\begin{subfigure}[t]{8.8cm}
\caption{}
\subcaptiontitle{CIFAR-10 replication a (30 subjects)}
\includegraphics[width=8.8cm]{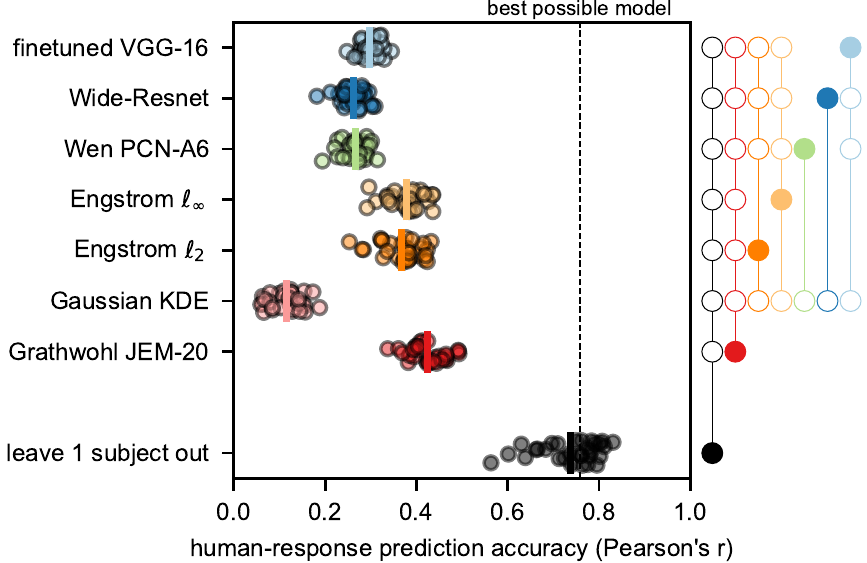}
\label{fig:CIFAR10_human_responses_prediction_mean_correlation_replication_a}
\end{subfigure}\hfill%
\begin{subfigure}[t]{8.8cm}
\caption{}
\subcaptiontitle{CIFAR-10 replication b (30 subjects)}
\includegraphics[width=8.8cm]{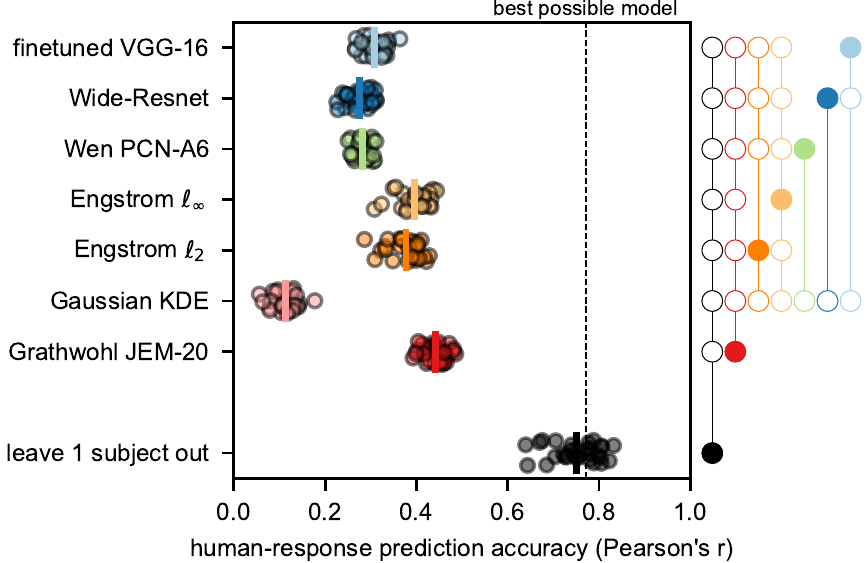}
\label{fig:CIFAR10_human_responses_prediction_mean_correlation_replication_b}
\end{subfigure}
\end{minipage}}
\caption{\textbf{A side by side comparison of two replications of Experiment 2.} Each replication was conducted with an independently synthesized set of 420 controversial stimuli and an independently selected set of 60 CIFAR-10 test examples. 30 subjects participated in each replication and no subject participated in both of them. Main text Fig.~\ref{fig:CIFAR-10_human_responses_prediction_mean_correlation} shows the same data pooled over the two replications. \textbf{The similarity of the outcomes of the two replications demonstrates the statistical reliability of our experimental procedure as a whole.}}
\label{fig:CIFAR10_human_responses_prediction_mean_correlation_two_replications}
\end{figure}

We ran a behavioral experiment similar to Experiment 1, presenting 420 controversial stimuli (20 per model-pair comparison) as well as 60 randomly selected CIFAR-10 test images. We ran two replications of the experiment on 30 subjects each, using a new, independent batch of controversial stimuli for each replication. The results pooled over both replications (60 subjects) are presented in Fig.~\ref{fig:CIFAR-10_human_responses_prediction_mean_correlation}, whereas the (virtually identical) results of each individual replication are presented in Fig.~\ref{fig:CIFAR10_human_responses_prediction_mean_correlation_two_replications}.

On average across the stimulus set, JEM-20 was significantly more accurate at predicting the human perceptual judgments than all other models. Similarly to Experiment 1, none of the models reached the lower bound of the noise ceiling (the leave-one-subject-out estimate). The two adversarially trained models (trained on $\ell_\infty$ and $\ell_2$ bounded perturbations) were second to the JEM-20 model in their human-response prediction accuracy. Next was the finetuned VGG-16 model, and then the discriminatively trained Wide-Resnet and the Predictive Coding Network. The Gaussian KDE had the lowest human-response prediction accuracy.

\begin{figure}%
\makebox[\textwidth][c]{\begin{minipage}{7in}%
\begin{subfigure}[t]{8.8cm}
\caption{}
\subcaptiontitle{MNIST, controversial stimuli}
\centering
\includegraphics[width=8.8cm]{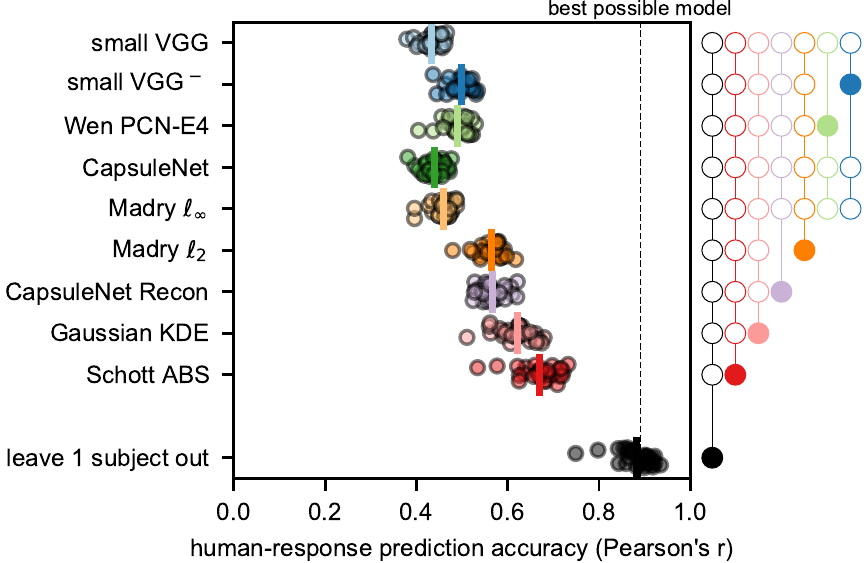}
\label{fig:MNIST_human_responses_prediction_mean_correlation_only_controvesial}
\end{subfigure}\hfill%
\begin{subfigure}[t]{8.8cm}
\caption{}
\subcaptiontitle{CIFAR-10, controversial stimuli}
\centering
\includegraphics[width=8.8cm]{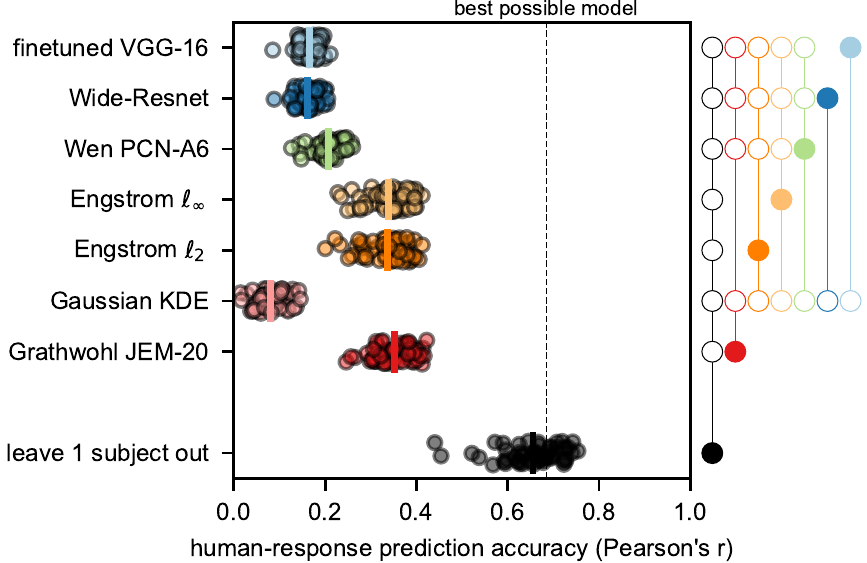}
\label{fig:CIFAR-10_human_responses_prediction_mean_correlation_only_controversial}
\end{subfigure}
\begin{subfigure}[t]{8.8cm}
\caption{}
\subcaptiontitle{MNIST, natural stimuli}
\centering
\includegraphics[width=8.8cm]{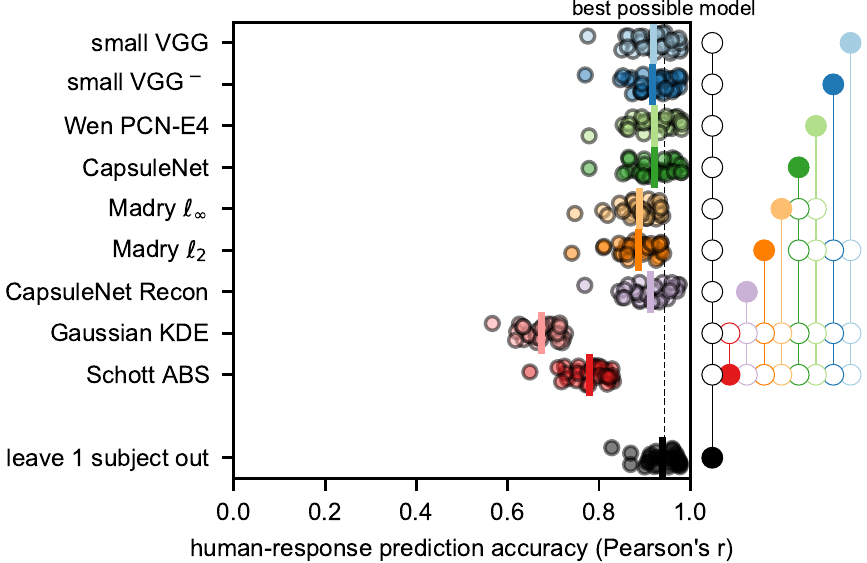}
\label{fig:MNIST_human_responses_prediction_mean_correlation_only_natural}
\end{subfigure}\hfill%
\begin{subfigure}[t]{8.8cm}
\caption{}
\subcaptiontitle{CIFAR-10, natural stimuli}
\centering
\includegraphics[width=8.8cm]{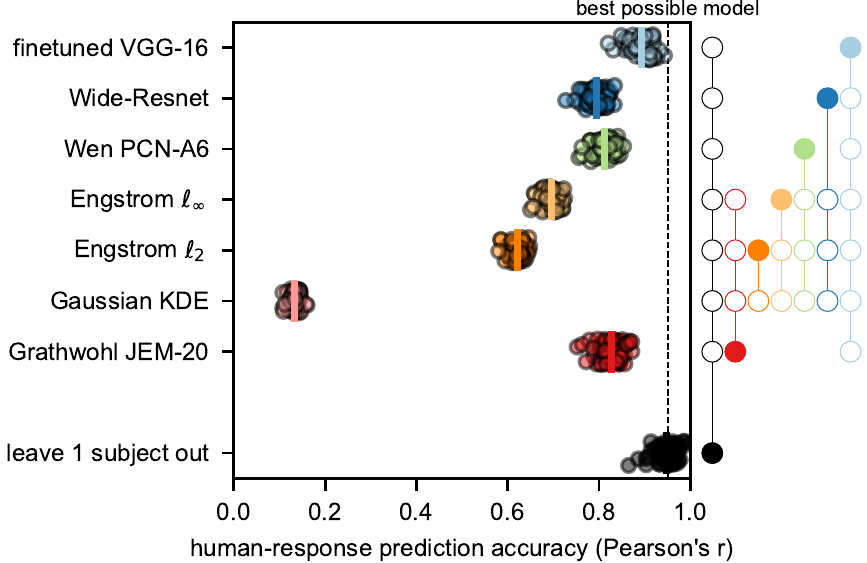}
\label{fig:CIFAR-10_human_responses_prediction_mean_correlation_only_natural}
\end{subfigure}\end{minipage}}%

\caption{\textbf{The performance of the candidate models in predicting the human responses to controversial stimuli (panel A, Experiment 1; panel B, Experiment 2) and to natural stimuli (panel C, Experiment 1; panel D, Experiment 2).} Unlike Fig.~\ref{fig:human_responses_prediction_mean_correlation} (main text), where the model-human response correlation coefficients were calculated across all of the stimuli included in the behavioral experiments, here the correlation coefficients were calculated separately for controversial and natural stimuli. The Schott ABS model \cite{schott_towards_2019}, which excels in the controversial stimuli benchmark (panel A), lags behind all of the discriminative MNIST models in predicting the responses to MNIST `natural' test examples (panel C). In contrast, the hybrid discriminative-generative JEM-20 \cite{grathwohl_your_2019} performs on par of its discriminative counterpart (`Wide-Resnet', employing the same architecture) in predicting the human response to test CIFAR-10 images (panel D). At the same time, JEM-20 is at least as good as the adversarially trained models (Engstrom $\ell_{\infty}$ and Engstrom $\ell_2$, \cite{engstrom_robustness_2019}) in predicting the human responses to the controversial stimuli (panel B). This relatively high performance both within and outside the training distribution demonstrates the advantage of a hybrid discriminative-generative modeling approach. However, JEM-20 is still less accurate than the finetuned, ImageNet-trained VGG-16 model in predicting the human responses to natural (test) CIFAR-10 images (panel D).}
\label{fig:prediction_accuracy_by_stimuli_type}
\end{figure}

Measuring the human response-prediction accuracy separately for controversial stimuli (Fig.~\ref{fig:CIFAR-10_human_responses_prediction_mean_correlation_only_controversial}) showed no significant difference between the JEM-20 model and the adversarially trained DNNs. For the natural images, however, the JEM-20 model significantly outperformed the adversarially trained DNNs (Fig.~\ref{fig:CIFAR-10_human_responses_prediction_mean_correlation_only_natural}). The model that best predicted the human responses to the natural images was the finetuned ImageNet-trained VGG-16, indicating that no single model in our candidate set was uniformly dominant, as would be expected of the true model.

\section*{Discussion}

In this paper, we introduce the method of synthetic controversial stimuli and we demonstrate its utility for adjudicating among DNNs as models of human recognition in the context of two simple visual recognition tasks, MNIST and CIFAR-10. Controversial stimuli reveal model differences and empower us to find failure modes, capitalizing on the fact that if two models disagree, at least one of them must be wrong.%

The method of controversial stimuli can be useful to two groups of scientists. The first group is cognitive computational neuroscientists interested in better understanding perceptual processes, such as object recognition, by modeling them as artificial neural networks. The second group is computer scientists interested in comparing the robustness of different DNN models to adversarial attacks.

\subsubsection*{Controversial stimuli offer a severe test for DNNs as brain-computational models}
Natural stimuli will always remain a necessary benchmark for models of perception. Scientists designing experiments can search for natural controversial stimuli to increase the power of model comparison. However, natural stimuli, even controversial ones, are insufficient for testing models of perception. In particular, a mechanistically incorrect model with many parameters that has been trained on natural images can achieve high performance at predicting human-assigned labels of images sampled from the same distribution. Synthetic controversial stimuli that are not limited to the training distribution provide a severe test of a model's inductive bias because they require the model to generalize far beyond the training distribution. Similarly, comparing models of perception using synthetic controversial stimuli ensures that we do not favor models that are higher capacity function approximators but are less functionally consistent with human perception.

\subsubsection*{Controversial stimuli generalize adversarial attacks}

Engineers use adversarial examples to test the robustness of models. Adversarial examples can be viewed as a special case of controversial stimuli. An ideal adversarial example is controversial between the targeted model and ground truth. In principle, therefore, adversarial examples require the evaluation of ground truth in the optimization loop. However, the evaluation of ground truth is often difficult, because it may be costly to compute or may require human judgment. In practice, adversarial attacks usually use a stand-in for ground truth, such as the assumption that the true label of an image does not change within a pixel-space $\ell_p$ ball of radius $\epsilon$.

Controversial stimulus synthesis enables us to compare two models in terms of their robustness without needing to evaluate or approximate the ground truth within the optimization loop. We only require a single ground-truth evaluation once the optimization is completed to determine which of the models responded incorrectly. Hence, controversial stimuli enable us to use more costly and compelling evaluations of ground truth (e.g., human judgments or a computationally complex evaluation function), instead of relying on a surrogate measure.

The most common surrogate measure for ground truth is $\epsilon$-robustness. A model is said to be $\epsilon$-robust if perturbations of the image confined to some distance in image space (defined by an $l_p$-norm) do not change the model's classification. The notion of $\epsilon$-robustness has led to analytical advances and enables adversarial training \cite{madry_towards_2018,ilyas_adversarial_2019}. However, since $\epsilon$-robustness is a simple surrogate for a more complicated ground truth, it does not preclude the existence of adversarial examples, and so does not guarantee robustness in a more general sense. This is particularly evident in the case of object recognition in images, where the ground truth is usually human categorization: A model can be $\epsilon$-robust for a large $\epsilon$ and yet show markedly human-inconsistent classifications, as demonstrated by controversial stimuli (here), distal adversaries \cite{schott_towards_2019}, and 'invariance attacks' \cite{jacobsen_exploiting_2019}, in which a human subject manually changes the true class of an image by making modifications confined to an $\ell_p$-ball in image space. The motivating assumption of $\epsilon$-robustness is that the decision regions are compact and their boundaries are far from the training examples. This does not hold in general. Controversial stimuli allow us to find failure modes in two or more models by studying differences in their decision boundaries instead of relying on assumptions about the decision boundaries.

\subsubsection*{Controversial stimuli: current limitations and future directions}

Like most works using pre-trained models \cite{kriegeskorte_deep_2015,yamins_using_2016}, this study operationalized each model as a single trained DNN instance. In this setting, a model predicts a single response pattern, which should be as similar as possible to the average human response. To the extent that the training of a model results in instances that make idiosyncratic predictions, the variability across instances will reduce the model's performance at predicting the human responses. An alternative approach to evaluating models would be to use multiple instances for each model \cite{mehrer_individual_2020}, considering each DNN instance as an equivalent of an individual human brain. In this setting, each model predicts a distribution of input-output mappings, which should be compared to the distribution of stimulus-response mappings across the human population. Instance-specific idiosyncrasies may then be found to be consistent (or not) with human idiosyncratic responses.

Another limitation of our current approach is scaling up: Synthesizing controversial stimuli for every pair of classes and every pair of models is difficult for problems with a large number of classes or models. A natural solution to this problem would be subsampling, where we do not synthesize the complete cross-product of class pairs and model pairs.%

Future research should also explore whether it is possible to replace the controversiality index with an optimal experimental design approach, jointly optimizing a stimulus set to reduce the entropy of our posterior over the models. Finally, adaptive measurement between or within experimental sessions could further increase the experimental efficiency.

\subsubsection*{Generative models may better capture human object recognition}
One interpretation of the advantage of the best performing models (the VAE-based Analysis By Synthesis model in Experiment 1 and the Joint Energy Model in Experiment 2) is that, like these two models, human object recognition includes elements of generative inference. There has recently been considerable progress with DNNs that can estimate complex image distributions (e.g., VAEs and normalizing-flow models). However, such approaches are rarely used in object recognition models, which are still almost always trained discriminatively to minimize classification error. Our direct testing of models against each other suggests that DNN classifiers that attempt to learn the distribution of images (in addition to being able to classify) provide better models of human object recognition.

However, none of the tested models approached the noise ceiling, and while the ABS and JEM models performed better than all of the other models on average, they were worse than some of the discriminative models when the natural examples were considered in isolation (Fig.~\ref{fig:prediction_accuracy_by_stimuli_type}C, D). Each of these two outcomes indicates that none of the models were functionally equivalent to the process that generated the human responses.

Generative models do not easily capture high-level, semantic properties of images \cite{nalisnick_deep_2018,fetaya_understanding_2020}. In particular, this problem is evident in the tendency of various deep generative models to assign high likelihood to out-of-distribution images that are close to the mean low-level statistics of the in-distribution images \cite{nalisnick_deep_2018}. Hybrid (discriminative-generative) approaches such as the joint energy model \cite{grathwohl_your_2019} are a promising middle-ground, yet the particular hybrid model we tested (JEM-20) was still far from predicting human responses accurately. An important challenge is to construct a generative or hybrid model that (a) reaches the noise ceiling in explaining human judgments, (b) scales up to real-world vision (e.g., ImageNet), and (c) is biologically plausible in both its architecture and training. The method of controversial stimuli will enable us to severely test such future models and resolve the question of whether human visual judgments indeed employ a process of generative inference, as suggested by our results here.

\section*{Materials and Methods}
Further details on 
training/adaptation of candidate models, stimulus optimization and selection, human testing and noise-ceiling estimation appear in the Supplementary Information.
\subsubsection*{Controversial stimuli synthesis}
Each controversial stimulus was initialized as a randomly seeded, uniform noise image ($x \sim \unif(0,1)$, where 0 and 1 are the image intensity limits). To efficiently optimize the controversiality score (Eq.~\ref{eq:C_AB}), we ascended the gradient of a more numerically favorable version of this quantity:
\begin{equation}\label{eq:C_AB_optimization}
\begin{split}
\tilde{c}_{A,B}^{y_a,y_b}(x)=\mathop{\boldsymbol{LSE^-_\alpha}}\big\{l_A(y_a\mid x),-l_A(y_b\mid x), l_B(y_b\mid x),-l_B(y_a\mid x)\big\},
\end{split}
\end{equation} where $\mathop{\boldsymbol{LSE^-_\alpha}}=-\log\sum_i{exp^{-\alpha x_i}}$ (an inverted LogSumExp, serving as a smooth-minimum), $\alpha$ is a hyperparameter that controls the LogSumExp smoothness (initially set to $1$), and $l_A(y \mid x)$ is the calibrated logit for class $y$ (the input to the sigmoid readout). Experiment-specific details on stimulus-optimization appear in the Supplementary Information sections \ref{MNIST_stimulus_synthesis details} and \ref{CIFAR10_stimulus_synthesis details}.

\subsubsection*{Human subjects}
90 participants took part in the online experiments and were recruited through prolific.co. All participants provided informed consent at the beginning of the study, and all procedures were approved by the Columbia Morningside ethics board.

\subsubsection*{Statistical inference}
Differences between models with respect to their human response prediction accuracy were tested by bootstrapping-based hypothesis testing. For each bootstrap sample (100,000 resamples), subjects and stimuli were both randomly resampled with replacement. Stimuli resampling was stratified by stimuli conditions (one condition per model pair, plus one condition of natural examples). For each pair of models $M_1$ and $M_2$, this bootstrapping procedure yielded an empirical sampling distribution of $r_{M_1}-r_{M_2}$, the difference between the models' prediction accuracy levels. Percent of bootstrapped accuracy differences below (or above) zero were used as left-tail (or right-tail) \textit{p}-values. These \textit{p}-values were Holm-Šídák corrected for multiple pairwise comparisons and for two-tailed testing.

\subsubsection*{Data and code availability}
Optimization source code, synthesized images, and detailed behavioral testing results will be available at \href{https://github.com/kriegeskorte-lab}{github.com/kriegeskorte-lab}.

\small{
\subsubsection*{Acknowledgments}
This material is based upon work supported by the National Science Foundation under Grant No. 1948004. TG acknowledges ELSC brain sciences postdoctoral fellowships for training abroad, and NVIDIA for a donation of a Titan Xp used for this research. 
The authors thank Máté Lengyel for a helpful discussion and Raphael Gerraty, Heiko Schütt, Ruben van Bergen, and Benjamin Peters for their comments on the manuscript.}
\bibliographystyle{unsrtnat} 
\bibliography{references}

\startsupplement{}
\newpage
\section*{Supplementary Materials and Methods}
\label{sec:supplementary_material_and_methods}

\subsection{Candidate MNIST models (Experiment 1)}
\label{MNIST model details}
Most of the tested models (Table~\ref{tab:MNIST_models_table}) were based on official pre-trained versions \cite{wen_deep_2018,sabour_dynamic_2017,madry_towards_2018,schott_towards_2019}, unmodified except for the readout layer. Here we describe the models we trained from scratch or more deeply altered.
\subsubsection{Small VGG} \label{Small VGG} Starting from the VGG-16 architecture \cite[architecture D in Table 1 of reference][]{simonyan_very_2014}, we downsized its input to the 28$\times$28 pixels MNIST format, removed the deepest three convolutional layers and replaced the three fully-connected layers with a single, 512-unit fully-connected layer, feeding a ten-sigmoid readout layer. All weights were initialized using the Glorot uniform initializer, as implemented in Keras. Batch normalization was applied between the convolution and the ReLU operations in all convolutional layers. The model was trained with Adagrad ($\textrm{learning rate}=10^{-3}$, $\epsilon=10^{-8}$, decay=0) for 20 epochs using a mini-batch size of 128. The epoch with best validation performance (evaluated on 5000 MNIST held-out training examples) was used.

\subsubsection{Reconstruction-based readout of the Capsule Network} In the training procedure of the original Capsule network \cite{sabour_dynamic_2017}, the informativeness of the class-specific activation vectors ('DigitCaps') is promoted by minimizing the reconstruction error of a decoder that is trained to read out the input image from the vector activation related to each example's correct class. \cite{frosst_darccc_2018,qin_detecting_2020} suggested to use this reconstruction error during inference, flagging examples with high reconstruction error (conditioned on their inferred class) as potentially adversarial. While rejecting suspicious images and avoiding their classification is a legitimate engineering solution, for a vision model we require that class conditional probabilities ($\hat{p}(y \mid x)$) will always be available. Hence, instead of using the reconstruction error as a rejection criterion, we used it as a classification signal. We modified the same official pre-trained Capsule Network (\href{https://github.com/Sarasra/models/tree/master/research/capsules}{https://github.com/Sarasra/models/tree/master/research/capsules}) used in our testing of the original Capsule Network such that for each image during inference, the decoder network produced ten class-specific input image reconstructions. The ten class-conditional mean squared reconstruction-errors were fed into ten sigmoids, whose response was calibrated as described in the results section. To eliminate the bias of this error measure towards blank images, we normalized the reconstruction error of each class by dividing it by the mean squared difference between the input image and the average image of all MNIST training examples (averaged across classes).
\subsubsection{Gaussian KDE}
\label{MNIST_Gaussian_KDE}
For each class $y$, we formed a Gaussian KDE model, $\hat{p}(x\mid y)=\frac{1}{n\sigma_y} \sum_{i=1}^n K\big(\frac{x-x^y_i}{\sigma_y}\big)$ where $\sigma_y$ is a class-specific bandwidth hyper-parameter, $K(\cdot)$ is a multivariate Gaussian likelihood with unit covariance, and $\{x^y_i\}$ are all MNIST training examples labeled as class $y$. $\sigma_y$ was chosen independently for each class from the range $[10^{-2},..,10^0]$ (100 logarithmic steps) to maximize the likelihood of held-out 500 training examples. The ten resulting log-likelihoods were fed as penultimate activations to a sigmoid readout layer, calibrated as described in the results section.

\subsection{Candidate CIFAR-10 models (Experiment 2)} \label{CIFAR-10 model details}
As in Experiment 1, most of the tested CIFAR-10 models (Table \ref{tab:CIFAR10_models_table}) were based on official pre-trained versions \cite{wen_deep_2018,engstrom_robustness_2019}, unmodified except for the readout layer. Models we trained from scratch or more deeply altered are described below.
\subsubsection{Finetuned VGG-16} \label{finetuned VGG-16} We initiated this model from TorchVision's \cite{marcel_torchvision_2010} batch-normalized implementation of VGG-16 \cite{simonyan_very_2014}, pretrained on ImageNet. We added a differentiable bilinear upsampling operation from $32 \times 32$ to $128 \times 128$ pixels before the first layer, replaced the DNN's final readout layer with ten sigmoids, one per CIFAR-10 class, and retrained the network on the CIFAR-10 training dataset to minimize the cross-entropy loss between the sigmoid outputs and CIFAR-10 one-hot labels for up to 100 epochs. We used stochastic gradient descent (PyTorch's implementation) with a learning rate of 0.001 and a momentum of 0.9. 5000 training examples (500 per class) were held out to serve as a validation dataset. We applied early stopping and selected the best epoch in terms of validation loss. 

\subsubsection{Gaussian KDE}
Fitting the pixel-space, class-specific CIFAR-10 Gaussian KDE followed the same procedure described for the MNIST Gaussian KDE (see SI subsection \ref{MNIST_Gaussian_KDE}), using a vectorized form of the $32\times32\times3$ RGB image representation as features. $\sigma_y$ was chosen independently for each class from the range $[10^{-2},..,10^2]$ (501 logarithmic steps).

\subsubsection{Joint Energy Model} 
We used the official code and pretrained model (\href{https://github.com/wgrathwohl/JEM}{https://github.com/wgrathwohl/JEM}) with two modifications: we increased the number of inference-time logit-refinement steps to 20 and reduced the stochasticity of this refinement by increasing the number of sampling chains from 5 to 30.
\subsubsection{Wide-Resnet}
\label{discriminative_wide_resnet_Training}
We trained the discriminative control of the JEM model ('Wide-Resnet') by executing the training code of JEM after adjusting the generative objective weight, '\verb|p_x_weight|', from 1.0 to 0.0. During inference, this model did not perform logit refinement.

\subsection{Synthesis of controversial stimuli: experiment-specific details}
\subsubsection{Controversial-stimulus synthesis in Experiment 1 (MNIST)} \label{MNIST_stimulus_synthesis details}

Each controversial stimulus was initialized to a floating-point $28\times28$ uniform white random noise matrix ($x \sim \unif(0.0,1.0)$, where 0.0 and 1.0 correspond to MNIST's 0 and 255 integer intensity levels, respectively). While for most candidate models included in Experiment 1, one can derive an analytical gradient of Eq.~\ref{eq:C_AB_optimization}, this is not possible for the ABS model since its inference is based on latent space optimization. Hence, following \cite{schott_towards_2019}'s approach to forming adversarial examples, we used numerical differentiation for all models. In each optimization iteration, we used the symmetric finite difference formula $\frac{f(x + h) - f(x-h)}{2h}$ to estimate the gradient of Eq.~\ref{eq:C_AB_optimization} with respect to the image. An indirect benefit of this approach is that one can set $h$ to be large, trading gradient precision for better handling rough cost-landscapes. For each image, we began optimizing using $h=1$ (clipping $(x+h)$ and $(x-h)$ to stay within the the grayscale intensity range). Once the optimization converged to a local maximum, we halved $h$ and continued optimizing. We kept halving $h$ upon convergence until final convergence with $h=1/256$. We then increased the LSE hyperparameter $\alpha$ to 10 and reset $h$ to equal 1.0 again, repeating the procedure (but without resetting the optimized image). A third and final optimization epoch used $\alpha=100$.

In each optimization iteration, once a gradient estimate was determined we used a line search for the most effective step size: We evaluated the effect of the maximal step in the direction of the gradient that did not cause intensity clipping, as well as $[2^{-1},2^{-2},...,2^{-8}]$ of this step size.

When the optimization converged to an image that had a controversiality score (Eq.~\ref{eq:C_AB}) of less than 0.85 we repeated the optimization procedure with a different initial random image, up to five attempts.

For analytically differentiable MNIST models, we found that this more involved (and more computationally intensive) approach to image optimization resulted in less convergence to poor local maxima (i.e., images with low controversiality) compared to standard gradient ascent using symbolic differentiation.

\subsubsection{Controversial-stimulus synthesis in Experiment 2 (CIFAR-10)} \label{CIFAR10_stimulus_synthesis details}
Since all of the candidate CIFAR-10 models were differentiable, we applied a symbolic-differentiation-based stimulus synthesis procedure in Experiment 2. Unlike the MNIST case, optimizing controversiality with the symbolic gradients of CIFAR-10 models rarely led to convergence to poor local maxima, potentially indicating a smoother cost landscape associated with the CIFAR-10 models.

The optimized image was parameterized as an unconstrained floating point 32$\times$32$\times$3 matrix $x_0$. This representation was transformed to an intensity image $x$ (with pixel intensities constrained between 0.0 and 1.0) by means of the sigmoid function, $x=\frac{1}{1+\exp(-x_0)}$. The image $x$ was fed to the target models, and eventually presented to the human subjects. $x$ was initilized as a uniform random noise image ($x \sim \unif(0.0,1.0)$) and the corresponding initial $x_0$ was set by the inverse sigmoid transform, $x_0=\log(\frac{x}{1-x})$.

Using \eqref{eq:C_AB_optimization} as our optimization objective, we applied Adam \cite{kingma_adam_2014} ($\alpha=0.1$, $\beta_1=0.9$, $\beta_2=0.999$, $\epsilon=10^{-8}$) to the unbounded image representation $x_0$. The smooth-minimum sharpness parameter $\alpha$ in \eqref{eq:C_AB_optimization} was initially set to 1.0. The optimizer was run to convergence and then resumed (without resetting the image) with a sharpness parameter of 10.0, and then finally with a sharpness parameter of 100.0. The convergence criterion was an improvement of less than 0.1\% in the maximal controversiality score in the last 50 time steps compared to the maximal controversiality score in the time steps that preceded this window. When the resulting image had a controversiality score (Eq.~\ref{eq:C_AB}) of less than 0.85, we repeated the optimization procedure with a different initial random image (up to five attempts). In most cases, even a single repetition was not needed.

\subsection{Selection of controversial stimuli for human testing}\label{controversial stimulus selection}
For each model pair, we selected 20 controversial stimuli for human testing (out of up to 90 we produced). Using integer programming (IBM DOcplex), we searched for the set of 20 images with the highest total controversiality score, under the constraint that each class is targeted exactly twice per model. This was done separately for Experiment 1 and Experiment 2.

\subsection{Human testing}
The two human experiments were conducted using a custom javascript interface. In addition to collecting the perceptual judgments, we monitored reaction times to detect too quick responses. Trials completed in less than 100 ms were rejected post hoc, treating the corresponding perceptual judgments as missing values. No participant took part in more than one experiment.

\subsubsection{Experiment 1}
30 participants (17 women, mean age = 29.3) participated in Experiment 1.

We monitored the participants' performance through three measures: their accuracy on the 100 MNIST images, their reaction times, and the reliability of their responses to 108 controversial images (3 per model pair) that were displayed again at the end of the experiment. While the participants' performance on these measures varied, we found no basis for rejecting the data produced by any participant due to evident low effort or negligence. 

\subsubsection{Experiment 2}
A total of 60 participants (25 women, mean age = 26.1) participated in the two replications of Experiment 2. 

\begin{figure}[tbhp]
\begin{subfigure}[t]{6.0cm}
\centering
\caption{}
\subcaptiontitle{Experiment 1, MNIST}
\label{fig:Experiment_1_human_exp_demo_trial}
\centering
\includegraphics[height=7cm,keepaspectratio]{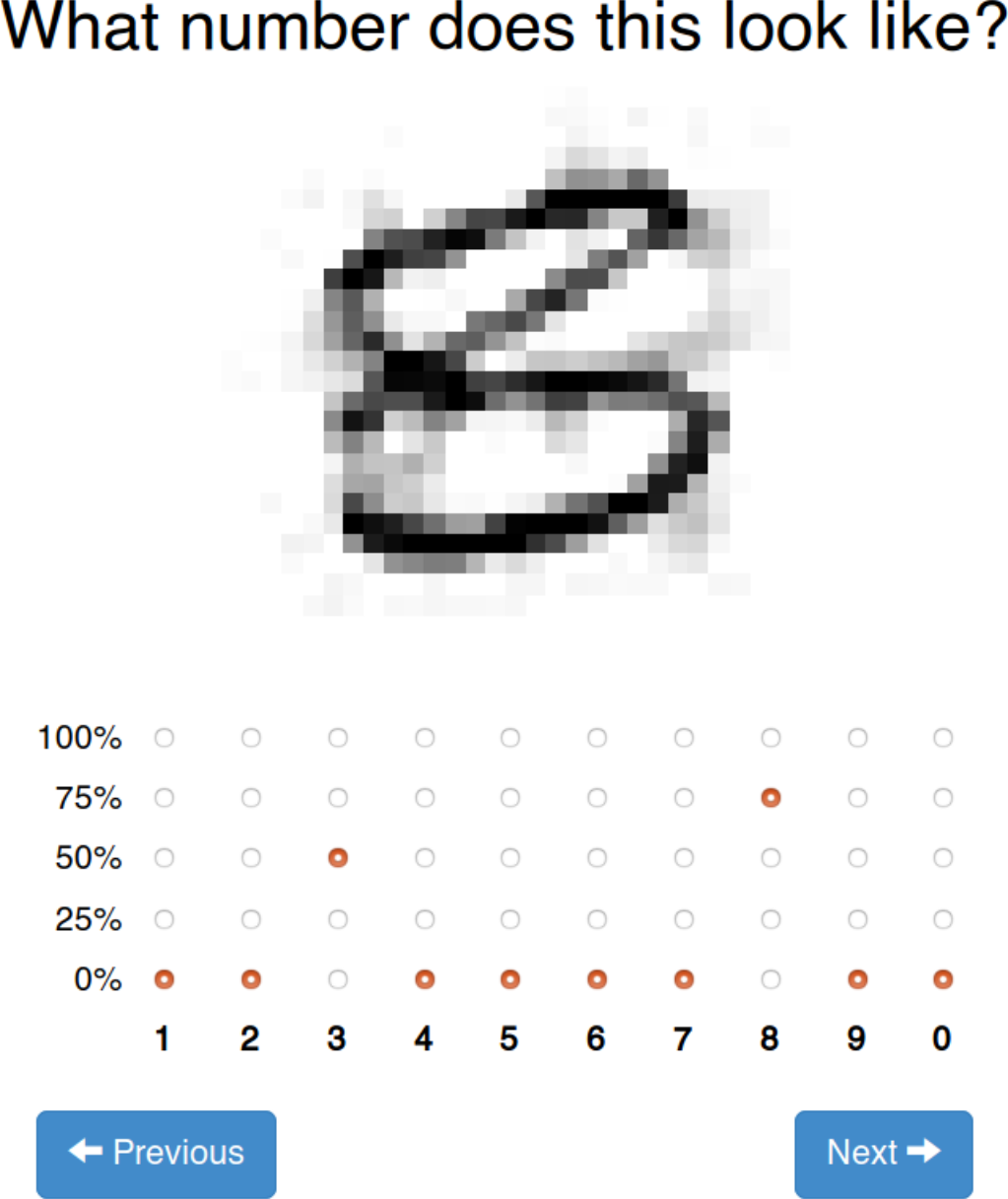}
\end{subfigure}\hfill%
\begin{subfigure}[t]{9.95cm}
\caption{}
\subcaptiontitle{Experiment 2, CIFAR-10}
\label{fig:Experiment_2_human_exp_demo_trial}
\centering
\includegraphics[height=7cm,keepaspectratio]{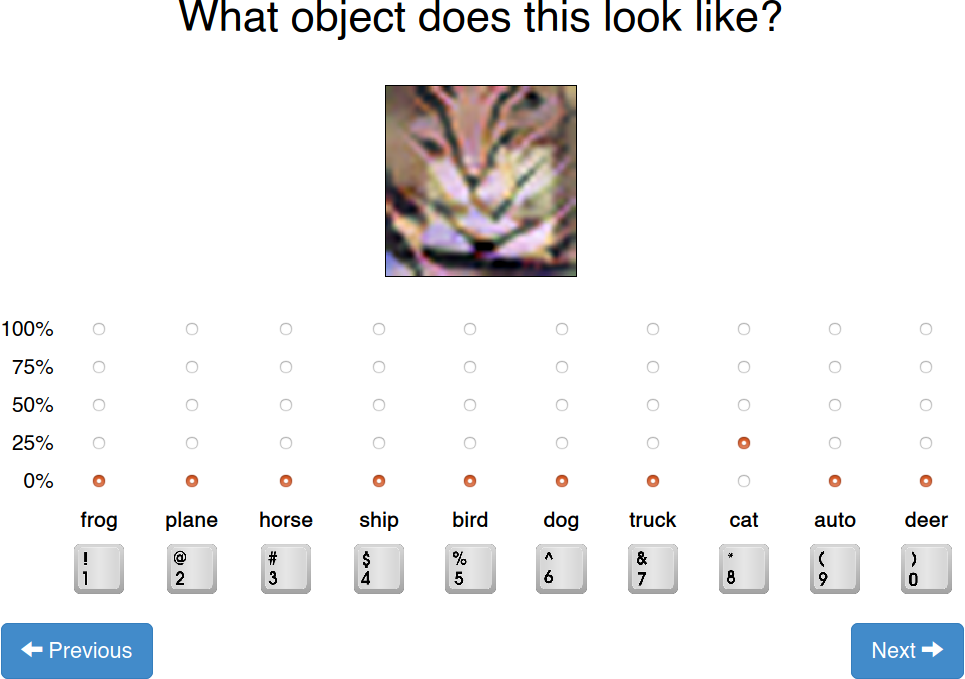}
\end{subfigure}\hfill%

\caption{\textbf{Trials in the human online experiments.} The subjects had to rate the presence of each class from 0\% to 100\%. These ratings do not need to sum to 100\%. Image presentation order was randomized for each subject. The 'Previous' button enabled subjects to go back one trial to correct their responses. (A) Experiment 1 (comparing MNIST models). The images were upsampled using nearest-neighbor interpolation. (B) Experiment 2 (comparing CIFAR-10 models). The images were upsampled using Lanczos interpolation. The mapping from the CIFAR-10 categories to the ten numerical response keys was randomized for each subject.}
\label{fig:human_exp_demo_trial}
\end{figure}

Since CIFAR-10 categories do not naturally map to response keys as MNIST categories do, we altered Experiment 1's graphical user interface to saliently display the mapping of categories to response keys (Fig.~\ref{fig:Experiment_2_human_exp_demo_trial}). This mapping was randomized for each participant.

As in Experiment 1, we monitored the performance of the human subjects through three measures: their accuracy on the 60 CIFAR-10 test images, their reaction times, and the reliability of their responses to 42 controversial images (2 per model pair) that were displayed again at the end of the experiment. The data of two participants were excluded due to suspected low effort performance (very fast completion time and all-zero ratings for natural several CIFAR-10 test examples). These two participants are not included in the total count of participating subjects above.

We found in pilot runs that a minority of the participants interpreted the task in an overly conservative way, assigning all-zero responses (i.e., no hint of an object) to all images that were not natural. We eliminated this kind of response pattern by including the following instruction:

\small\texttt{In each of the images that you will see, there will be hints for at least one of the ten object\\categories. If you see anything that reminds one of the ten object categories, rate the relevant\\category with at least 25\%. We will give a 5 USD worth bonus payment to any participant who will\\do well in detecting objects that are especially hard to recognize.}

\normalsize

This bonus was paid after the experiment to half of the participants according to an objective criterion. Since the bonus was awarded offline, it did not serve as feedback (which was intentionally absent in both experiments).

\subsection{Noise-ceiling estimates}
\label{noise_ceiling_details}
\subsubsection{Lower bound (leave-one-subject-out)}
Here we further detail the calculation of the leave-one-subject-out estimate, which serves as a lower bound on the noise ceiling (i.e., the true model should be at least as accurate as this estimate). To calculate this estimate, we held out one subject $s_i$ at a time and averaged the response patterns of all of the other subjects: If $\hat{p}_{s_i}(y\mid x)$ is the probability judgment provided by subject $s_i$ for image $x$ and category $y$, then the leave-one-subject-out prediction for this subject, image and category is given by $\hat{q}_{s_i}(y\mid x)=\frac{1}{n-1}\sum\limits_{j\neq i}\hat{p}_{s_j}(y\mid x)$. Considering all stimuli and categories, one can represent subject $s_i$'s response pattern and the corresponding leave-one-subject-out predicted response pattern as two vectors of matching lengths, $\bm{p}_{s_i}$ and $\bm{q}_{s_i}$. These vectors would have 8200 elements for the responses of an Experiment 1 subject (820 images $\times$ 10 response categories) or 4800 elements for the responses of an Experiment 2 subject (480 images $\times$ 10 response categories). The linear correlation between $\bm{p}_{s_i}$ and $\bm{q}_{s_i}$ measures how well can the subject's response pattern be predicted from the mean response pattern of her/his peers: 

\begin{equation}\label{eq:r_Pearson_leave_one_subject_out}
r(S,s_i)=\frac{
\sum\limits_{x,y}\big(\hat{p}_{s_i}(y\mid x) - \bar{\hat{p}}_{s_i}\big)
\big(\hat{q}_{s_i}(y\mid x) - \bar{\hat{q}}_{s_i}\big)}
{
\sqrt{\sum\limits_{x,y}\big(\hat{p}_{s_i}(y\mid x) - \bar{\hat{p}}_{s_i}\big)^2}
\sqrt{
\sum\limits_{x,y}\big(\hat{q}_{s_i}(y\mid x) - \bar{\hat{q}}_{s_i}\big)^2}
},
\end{equation}
where $\bar{\hat{p}}_{s_i}$ is the average probability judgment of subject $s_i$ and $\bar{\hat{q}}_{s_i}$ is the average probability judgment of the corresponding leave-one-subject-out predicted response pattern.

The resulting leave-one-subject-out correlation coefficients for each held-out subject were plotted as gray dots at the bottom of Fig.~\ref{fig:MNIST_human_responses_prediction_mean_correlation} and Fig.~\ref{fig:CIFAR-10_human_responses_prediction_mean_correlation}. The mean leave-one-subject correlation coefficient (averaged across subjects) was marked as a vertical bar in these two figures and was statistically tested against the mean model-human correlation coefficients of each candidate model. For analyses that included model recalibration, an inverse sigmoid transform was applied to the $\bm{q}_{s_i}$ vectors to produce logits, which were then tuned in the same fashion the models' logits were recalibrated, sharing the same scale and shift parameters across different held-out subjects, exactly matching the level of flexibility in fitting each model's predictions to the human data by scaling and shifting the model's logits.

\subsubsection{Upper bound ('best possible model')}
The upper bound on the noise ceiling (marked as 'best possible model' in Fig.~\ref{fig:MNIST_human_responses_prediction_mean_correlation} and Fig.~\ref{fig:CIFAR-10_human_responses_prediction_mean_correlation}) was determined by optimization. We initiated a vector of logits (one value per image-category combination, e.g., 8200 elements for Experiment 1, or 4800 elements for Experiment 2) as a zero vector and optimized this vector with L-BFGS so that the across-subject mean of the correlation coefficients between each subject and a sigmoid transform of this vector is maximized. The resulting mean correlation coefficient reflects the inherent limitation of predicting variable individual response patterns by a single response pattern.

Note that for the simple case of no missing values, the vector that maximizes the across-subject average correlation is directly obtainable by transforming each individual response vector to a z-scored vector and then averaging the resulting z-scored vectors across subjects \cite[][supplementary materials]{nili_toolbox_2014}. Here, we used the more general optimization approach since the rejection of trials with too short reaction times led to missing values which complicate the analytical derivation of the correlation-maximizing response-vector.
\subsection{Additional software tools used} TensorFlow \cite{martin_abadi_tensorflow_2015} (Experiment 1), Keras \cite{chollet_keras_2015} (Experiment 1), and PyTorch \cite{paszke_pytorch_2019} (Experiments 1 \& 2) were used for DNN training and testing; psiTurk \cite{mcdonnell_psiturk_2012} was used for as the backend of the online experiments; Numpy \cite{walt_numpy_2011}, XArray \cite{hoyer_xarray_2017}, pandas \cite{mckinney_data_2010}, scikit-learn \cite{pedregosa_scikit-learn:_2011}, and statsmodels \cite{seabold_statsmodels_2010} were used for data analysis; matplotlib \cite{hunter_matplotlib_2007} and seaborn \cite{waskom_seaborn_2014} were used for visualization.

\newpage

\begin{figure}[h]
\centering
\includegraphics[width=\textwidth]{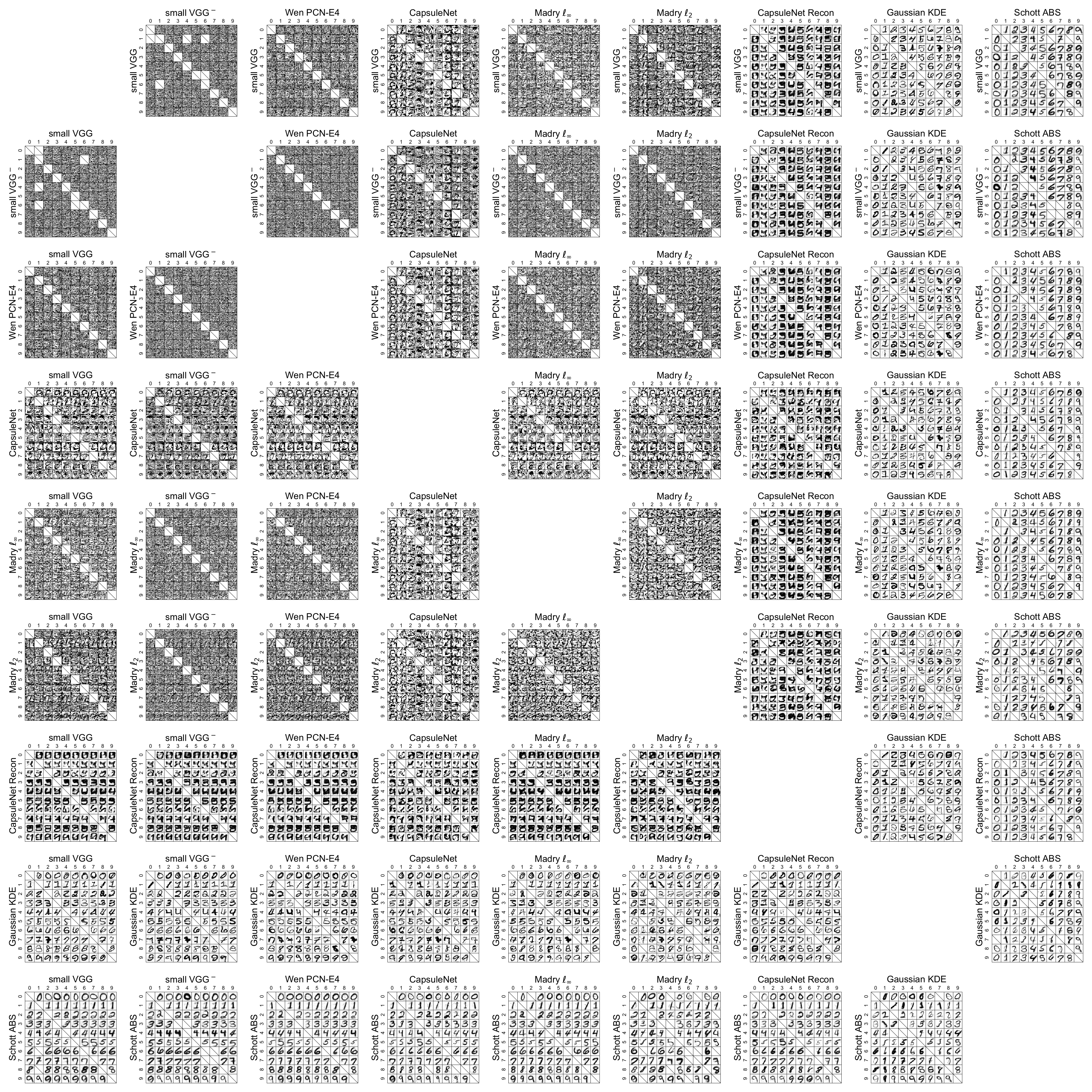}

\caption{\textbf{The entire set of controversial stimuli we synthesized for Experiment 1 (MNIST), organized by model pairs.} Each panel indicates a targeted model pair and the rows and columns within each panel indicate the targeted class pairs. For example, consider the bottom-left image in the bottom-left panel. This image (seen to us as a 9) is detected as a 0 by the small VGG model and as a 9 by the Schott ABS model. Missing (crossed) cells are either along the diagonal (where the two models would agree) or where our optimization procedure did not converge to a sufficiently controversial image (a controversiality score of at least 0.75). Best viewed digitally.}
\label{fig:MNIST_all_controversial_images_by_model_pairs}
\end{figure}

\begin{figure}%
\makebox[\textwidth][c]{\begin{minipage}{7in}%
\begin{subfigure}[t]{8.8cm}
\caption{}
\subcaptiontitle{MNIST, linear correlation coefficients,\\
with model recalibration}
\includegraphics[width=8.8cm]{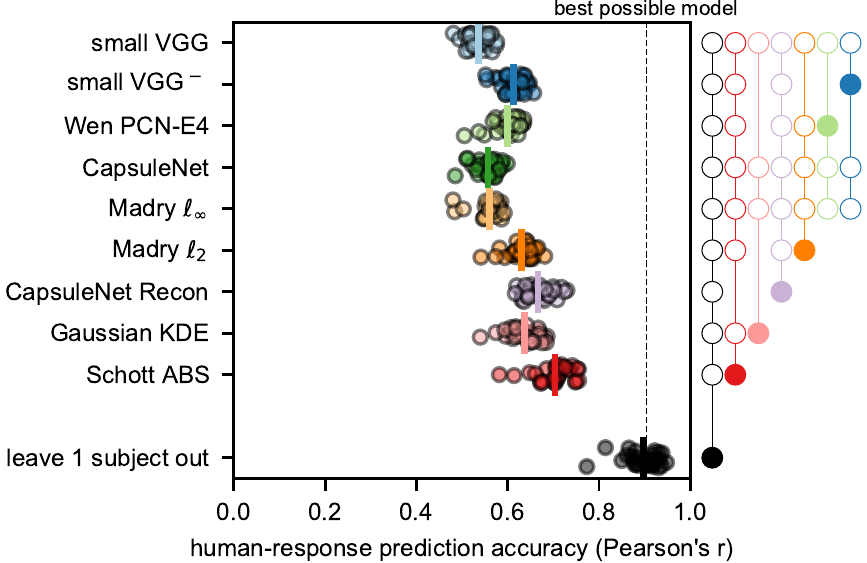}
\label{fig:MNIST_human_responses_prediction_mean_correlation_recalibrated}
\end{subfigure}\hfill%
\begin{subfigure}[t]{8.8cm}
\caption{}
\subcaptiontitle{CIFAR-10, linear correlation coefficients,\\ with model recalibration}
\includegraphics[width=8.8cm]{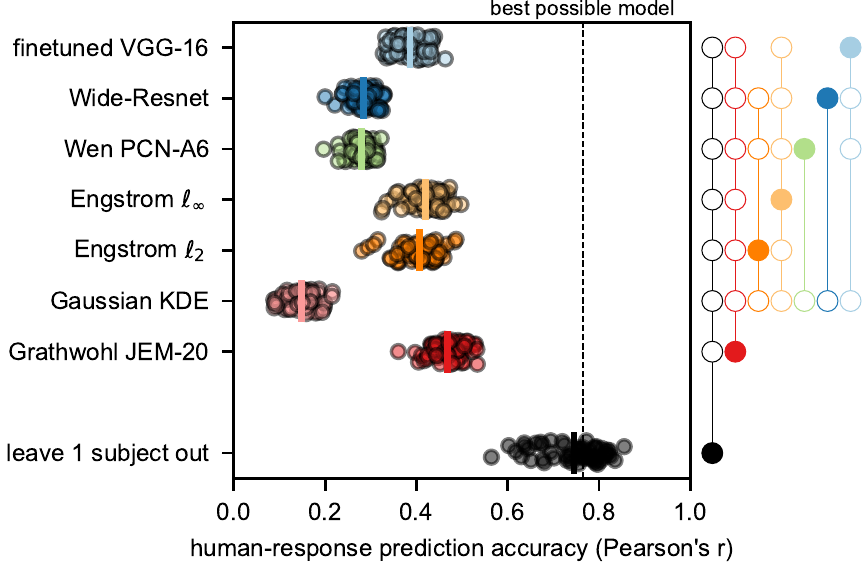}
\label{fig:CIFAR-10_human_responses_prediction_mean_correlation_recalibrated}
\end{subfigure}
\begin{subfigure}[t]{8.8cm}
\caption{}
\subcaptiontitle{MNIST, Mean Squared Error, with model recalibration}
\includegraphics[width=8.8cm]{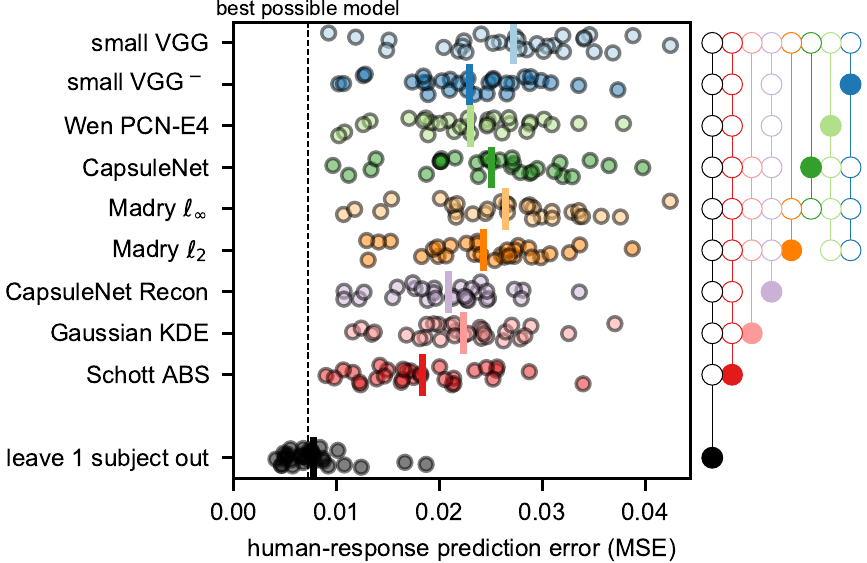}
\label{fig:MNIST_human_responses_prediction_MSE_recalibrated}
\end{subfigure}\hfill%
\begin{subfigure}[t]{8.8cm}
\caption{}
\subcaptiontitle{CIFAR-10, Mean Squared Error, with model recalibration}
\includegraphics[width=8.8cm]{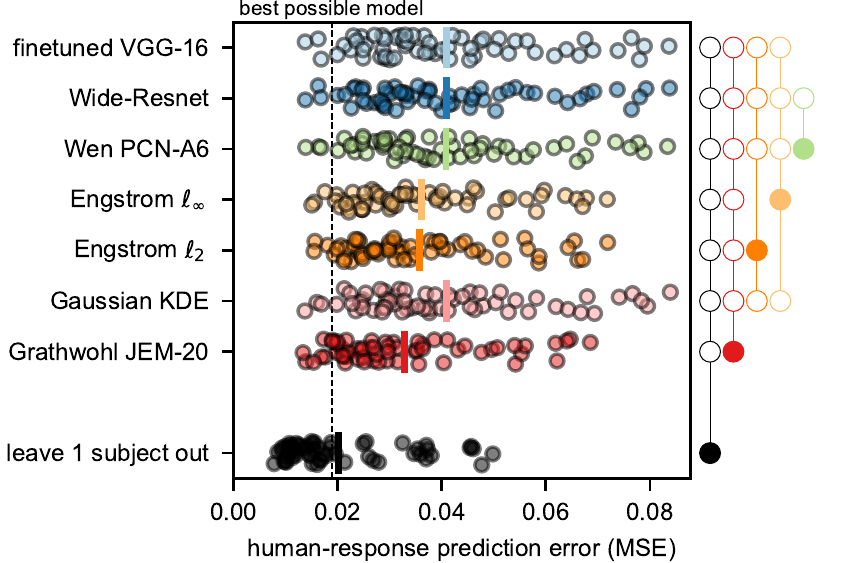}
\label{fig:CIFAR-10_human_responses_prediction_MSE_recalibrated}
\end{subfigure}\end{minipage}}

\caption{\textbf{Alternative measures of the model-human prediction accuracy show a similar rank ordering of models.} (A) Linear correlation coefficients between each candidate MNIST models and the human responses to all of the test stimuli (Experiment 1), after recalibrating each model independently to maximize its mean-across-subjects human response prediction accuracy. The logits of each model (i.e., the inputs to the ten readout sigmoids, see main text) were recalibrated by a linear transformation. The slope and intercept of this transformation were adjusted to maximize each model's mean correlation coefficient. The slope and intercept parameters were unique to each model but did not vary across classes and subjects (i.e., there were exactly two free parameters per model). Note that this procedure may introduce a small optimistic bias to all of the correlation measures. (B) An equivalent analysis of the CIFAR-10 models' predictions (Experiment 2). (C) Measuring Mean Squared Error (MSE) instead of linear correlation for the MNIST candidate models. Model recalibration was applied here as well, fitting the slope and intercept of each model's transformed logits to minimize the grand-average MSE across subjects. Unlike the linear correlation measure, the MSE measure does not reduce individual differences in the baseline and scale of human behavioral ratings. Hence the greater dispersion of the subjects' MSEs compared to the subjects' correlation coefficients. (D) An equivalent analysis for the CIFAR-10 candidate models.\\
The MSE between a model $M$ and subject $S_i$ was calculated as 
$\frac{1}{|X| |C|}\sum\limits_{x,y}\big(\hat{p}_{s_i}(y\mid x) - \hat{p}_M(y \mid x)\big)^2$ where $\hat{p}_{s_i}(y \mid x)$ is the human-judged probability that image $x$ contains class $y$, $\hat{p}_M(y \mid x)$ is the model's corresponding recalibrated judgment, $|X|$ is the number of stimuli (720 for Experiment 1, 480 for Experiment 2) and $|C|$ is the number of classes (ten for both experiments).
}%
\label{fig:alternative_model_human_prediction_accuracy_measures}
\end{figure}

\begin{figure}
\centering
\includegraphics[width=\textwidth]{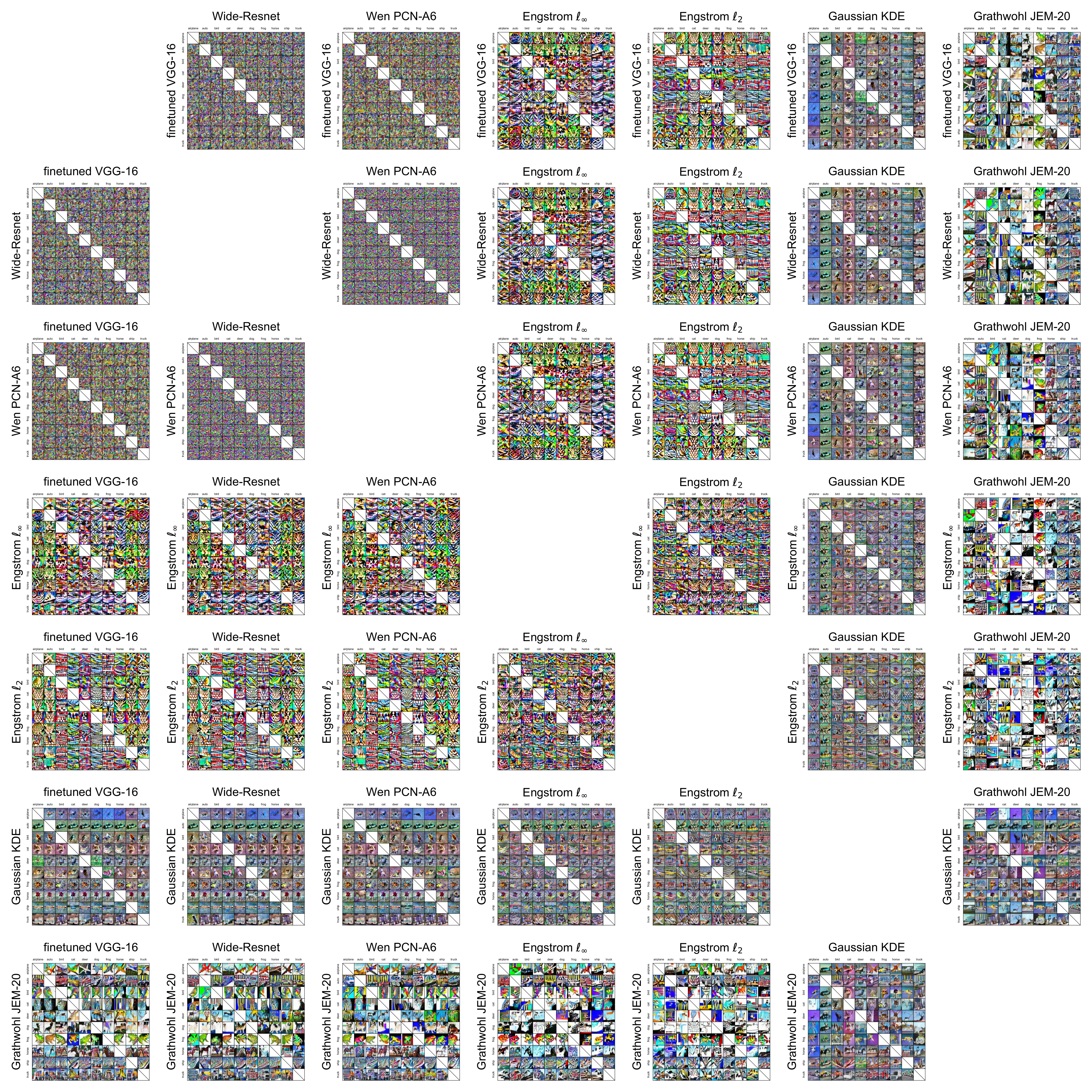}

\caption{\textbf{The entire set of controversial stimuli we synthesized for Experiment 2 (CIFAR-10), organized by model pairs.} Each panel indicates a targeted model pair, and the rows and columns within each panel indicate the targeted class pairs. For example, consider the bottom-left image in the bottom-left panel. This image is classified as an airplane by the finetuned VGG-16 model and as a truck by the Grathwohl JEM-20 model. Missing (crossed) cells are either along the diagonal (where the two models would agree) or where our optimization procedure did not converge to a sufficiently controversial image (a controversiality score of at least 0.75). Best viewed digitally.}
\label{fig:CIFAR10_all_controversial_images_by_model_pairs}
\end{figure}

\FloatBarrier
\renewcommand{\arraystretch}{1.1}
\begin{table}
            \caption{Candidate MNIST models}\label{tab:MNIST_models_table}
            \small
            \begin{tabular}{llcccc}

 & & & \multicolumn{3}{c}{human-response prediction accuracy}\\
model family & model & {MNIST test error} & all stimuli & controversial & natural \\
\midrule
\multirow{2}{*}{discriminative feedforward} & small VGG \cite{simonyan_very_2014}$^*$ &            0.47\% &                              0.516 &         0.434 &   0.919 \\
           & small VGG$^-$ \cite{simonyan_very_2014}$^*$ &            0.59\% &                              0.592 &         0.498 &   0.918 \\
\hline
\multirow{2}{*}{discriminative recurrent} & Wen PCN-E4 \cite{wen_deep_2018} &            0.42\% &                              0.567 &         0.490 &   0.921 \\
           & CapsuleNet \cite{sabour_dynamic_2017} &            0.24\% &                              0.541 &         0.439 &   0.921 \\
\hline
\multirow{2}{*}{adversarially trained} & Madry $\ell_\infty$ \cite{madry_towards_2018} \begin{scriptsize}$(\epsilon=0.3)$\end{scriptsize} &            1.47\% &                              0.538 &         0.459 &   0.890 \\
           & Madry $\ell_{2}$ \cite{madry_towards_2018} \begin{scriptsize}$(\epsilon=2.0)$\end{scriptsize} &            1.07\% &                              0.623 &         0.565 &   0.887 \\
\hline
\multirow{1}{*}reconstruction-based & CapsuleNet Recon \cite{qin_detecting_2020}$^*$ &            0.29\% &                              0.643 &         0.566 &   0.912 \\
\hline
\multirow{2}{*}{generative} & Gaussian KDE &            3.21\% &                              0.632 &         0.621 &   0.675 \\
           & Schott ABS \cite{schott_towards_2019} &            1.00\% &                              0.699 &         0.671 &   0.779 \\
\bottomrule
\end{tabular}
\\
\begin{adjustwidth}{0in}{0.65in}
\addtabletext{MNIST models tested in Experiment 1. 'Human-response prediction accuracy' is the across-subject average of the correlation between model and human judgments ($\bar{r}_{M}$). The values listed here correspond to the means depicted by vertical bars in Figs.~\ref{fig:MNIST_human_responses_prediction_mean_correlation},~\ref{fig:MNIST_human_responses_prediction_mean_correlation_only_controvesial}, and \ref{fig:MNIST_human_responses_prediction_mean_correlation_only_natural}.\\$*$ A modified architecture. See SI section \ref{MNIST model details}.}
\end{adjustwidth}
\end{table}

\begin{table}
            \caption{Canidate CIFAR-10 models}\label{tab:CIFAR10_models_table}
            \small
            \begin{tabular}{llllll}
 & & & \multicolumn{3}{l}{human-response prediction accuracy} \\
model family & model & CIFAR-10 test error & all stimuli & controversial & natural\\
\midrule
\multirow{2}{*}{discriminative feedforward} & finetuned VGG-16 \cite{simonyan_very_2014}$^*$ &               4.37\% &                              0.302 &         0.166 &   0.894 \\
                                   & Wide-Resnet \cite{grathwohl_your_2019} &               6.22\% &                              0.268 &         0.162 &   0.795 \\
\hline
discriminative recurrent & Wen PCN-A6 \cite{wen_deep_2018} &               6.66\% &                              0.274 &         0.208 &   0.812 \\
\hline
\multirow{2}{*}{adversarially trained} & Engstrom $\ell_\infty$ \cite{engstrom_robustness_2019} \begin{scriptsize}$(\epsilon=8/255)$\end{scriptsize} &              12.97\% &                              0.387 &         0.340 &   0.696 \\
                                   & Engstrom $\ell_{2}$ \cite{engstrom_robustness_2019} \begin{scriptsize}$(\epsilon=1.0)$\end{scriptsize} &              18.38\% &                              0.373 &         0.337 &   0.622 \\
\hline
\multirow{1}{*}{generative} & Gaussian KDE &              61.21\% &                              0.114 &         0.080 &   0.133 \\
\hline
\multirow{1}{*}{hybrid discriminative-generative} & Grathwohl JEM-20 \cite{grathwohl_your_2019} &               9.71\% &                              0.434 &         0.351 &   0.827 \\
\bottomrule
\end{tabular}
\\
\begin{adjustwidth}{0in}{0.0in}
\addtabletext{CIFAR-10 models tested in Experiment 2. 'Human-response prediction accuracy' is the across-subject average of the correlation between model and human judgments ($\bar{r}_{M}$). The values listed here correspond to the means depicted by vertical bars in Figs.~\ref{fig:CIFAR-10_human_responses_prediction_mean_correlation},~\ref{fig:CIFAR-10_human_responses_prediction_mean_correlation_only_controversial}, and \ref{fig:CIFAR-10_human_responses_prediction_mean_correlation_only_natural}.\\$*$ A modified architecture. See SI subsection \ref{finetuned VGG-16}.\\$\dagger$ Both models share the Wide-Resnet WRN-28-10 architecture \cite{zagoruyko_wide_2016}, trained without batch normalization, and differ in their training and inference procedures (see SI subsection \ref{discriminative_wide_resnet_Training}).}
\end{adjustwidth}
\end{table}

\end{document}